\documentclass[article]{IEEEtran}
\usepackage{times}
\usepackage[numbers]{natbib}
\usepackage{multicol}
\usepackage{graphicx}
\usepackage[bookmarks=true]{hyperref}
\usepackage[usenames, dvipsnames,table,xcdraw]{xcolor}
\usepackage{amsmath}
\usepackage[capitalise]{cleveref}
\usepackage{tabularx}
\usepackage{siunitx}

\usepackage{array}
\usepackage{makecell}
\usepackage{booktabs}
\usepackage{bbm}
\usepackage{caption}
\usepackage{subcaption}
\usepackage{authblk}
\usepackage{balance}

\makeatletter
\renewcommand\AB@affilsepx{, \protect\Affilfont}
\makeatother

\begin{document}

\title{Barkour: Benchmarking Animal-level Agility with Quadruped Robots}

\author{Ken Caluwaerts$^*$\thanks{* Equal contribution}}
\author{Atil Iscen$^*$}
\author{J. Chase Kew$^*$}
\author{Wenhao Yu$^*$}
\author{Tingnan Zhang$^*$}
\author{Daniel Freeman$^\dagger$\thanks{$\dagger$ Core contributor}}
\author{Kuang-Huei~Lee$^\dagger$}
\author{Lisa~Lee$^\dagger$}
\author{Stefano~Saliceti$^\dagger$}
\author{Vincent~Zhuang$^\dagger$}
\author{Nathan~Batchelor}
\author{Steven~Bohez}
\author{Federico~Casarini}
\author{Jose Enrique~Chen}
\author{Omar~Cortes}
\author{Erwin~Coumans$^\mathsection$\thanks{$\mathsection$ Work done while at Google.}}
\author{Adil~Dostmohamed}
\author{Gabriel~Dulac-Arnold}
\author{Alejandro~Escontrela$^\mathsection$}
\author{Erik~Frey}
\author{Roland~Hafner}
\author{Deepali~Jain}
\author{Bauyrjan~Jyenis}
\author{Yuheng~Kuang}
\author{Edward~Lee}
\author{Linda~Luu}
\author{Ofir~Nachum}
\author{Ken~Oslund}
\author{Jason~Powell}
\author{Diego~Reyes}
\author{Francesco~Romano}
\author{Feresteh~Sadeghi}
\author{Ron~Sloat}
\author{Baruch~Tabanpour}
\author{Daniel~Zheng}
\author{Michael~Neunert}
\author{Raia~Hadsell}
\author{Nicolas~Heess}
\author{Francesco~Nori}
\author{Jeff~Seto}
\author{Carolina~Parada}
\author{Vikas~Sindhwani}
\author{Vincent~Vanhoucke}
\author{Jie~Tan}
\affil{Google DeepMind}

\maketitle

\begin{abstract}

Animals have evolved various agile locomotion strategies, such as sprinting, leaping, and jumping. There is a growing interest in developing legged robots that move like their biological counterparts and show various agile skills to navigate complex environments quickly. Despite the interest, the field lacks systematic benchmarks to measure the performance of control policies and hardware in agility. We introduce the Barkour benchmark, an obstacle course to quantify agility for legged robots.  Inspired by dog agility competitions, it consists of diverse obstacles and a time based scoring mechanism. This encourages researchers to develop controllers that not only move fast, but do so in a controllable and versatile way. To set strong baselines, we present two methods for tackling the benchmark. In the first approach, we train specialist locomotion skills using on-policy reinforcement learning methods and combine them with a high-level navigation controller. In the second approach, we distill the specialist skills into a Transformer-based generalist locomotion policy, named Locomotion-Transformer, that can handle various terrains and adjust the robot's gait based on the perceived environment and robot states. Using a custom-built quadruped robot, we demonstrate that our method can complete the course at half the speed of a dog\footnote{See \href{https://sites.google.com/view/barkour}{sites.google.com/view/barkour} for supplementary materials.}. We hope that our work represents a step towards creating controllers that enable robots to reach animal-level agility.

\end{abstract}

\section{Introduction}
\label{sec:intro}

\begin{figure}[h!]
    \centering
    \begin{subfigure}{0.48\textwidth}
    \includegraphics[width=\textwidth]{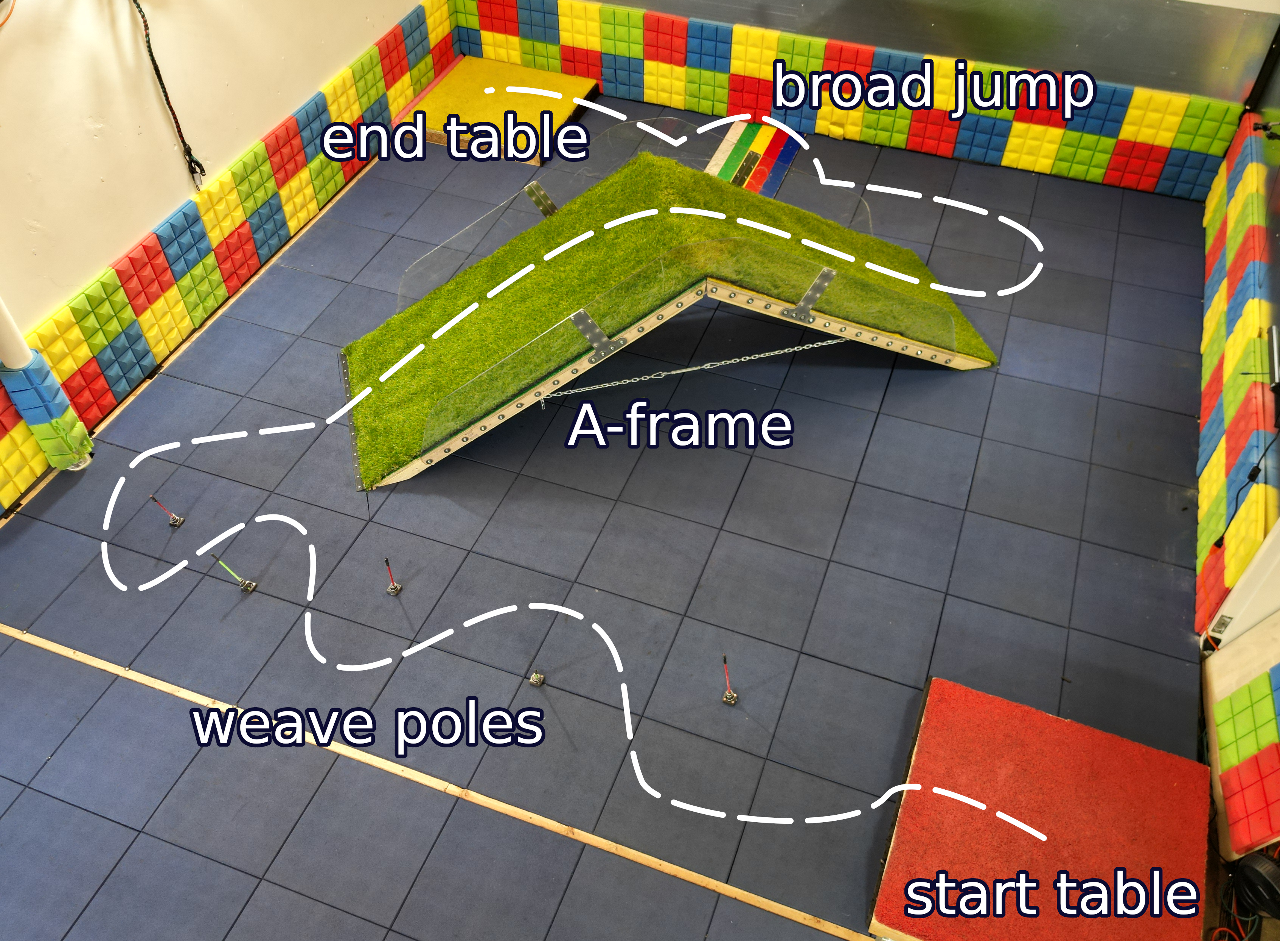}
    \caption{Physical setup (\SI{5}{\meter} x \SI{5}{\meter} course).}
    \vspace{3pt}
    \end{subfigure}
    \begin{subfigure}{0.48\textwidth}
    \includegraphics[width=\textwidth]{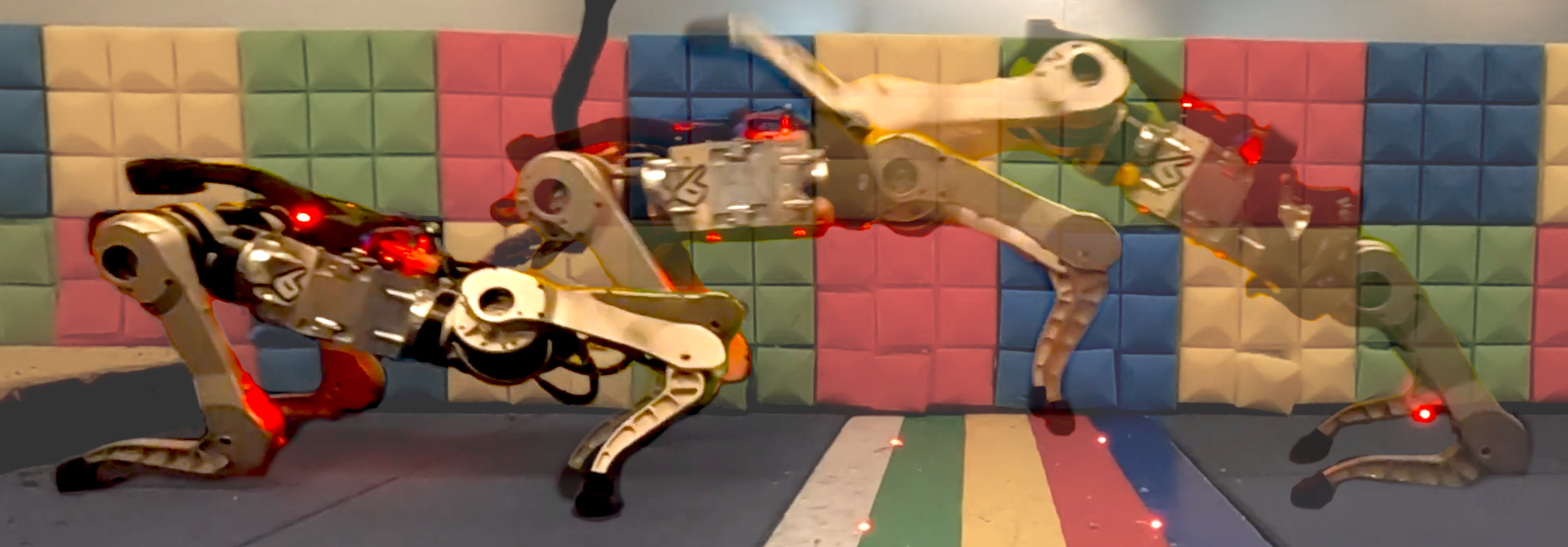}
    \caption{Typical behavior on \SI{0.5}{\meter} broad jump.}
    \end{subfigure}
    \caption{Barkour benchmark overview and example behavior.}
    \label{fig:teaser}
\end{figure}

There has been a proliferation of legged robot development inspired by animal mobility. Recent notable examples include the ETH ANYmal~\cite{hutter2016anymal}, the MIT Mini Cheetah~\cite{katz2019mini}, the KAIST RaiBo~\cite{choi2023learning}, Unitree A1/Go1, and the Boston Dynamics Spot robots. An important research question in this field is how to develop a controller that enables legged robots to exhibit animal-level agility while also being able to generalize across various obstacles and terrains. Through the exploration of both learning and traditional control-based methods, there has been significant progress in enabling robots to walk across a wide range of terrains~\cite{dicarlo2018, lee2020learning, kumar2021rma, agarwal2022legged, miki2022learning}. These robots are now capable of walking in a variety of indoor and outdoor environments, such as up and down stairs, through bushes, and over unpaved roads and rocky or even sandy beaches.

Despite advances in robot hardware and control, a major challenge in the field is the lack of standardized and intuitive methods for evaluating the effectiveness of locomotion controllers. Ad-hoc metrics are often used to present results, which complicates the comparing of results. To address this issue, it is essential to establish metrics that can accurately measure robot agility and to define a standard set of tasks that can serve as a common evaluation framework, similar to how the DeepMind Control Suite~\cite{tassa2018deepmind} has been widely adopted in the field of reinforcement learning (RL).

A good benchmark set for agile legged locomotion should (1) be non-trivial or not easily exploitable, and (2) require a diverse set of primitive behaviors that quadrupeds showcase in real environments. 
Well-established benchmarks already exist to measure animal performance, for example, dog agility competitions~\cite{akc2023}. In these competitions, participants race their dogs through a pre-set obstacle course. A variety of obstacles including weave poles, jumps, tunnels, an A-frame, a seesaw, and a pause table test diverse locomotion skills. The performance is evaluated based on time, and there are penalties for errors such as completing obstacles in the wrong order, tackling an obstacle from the wrong direction, or touching the jump bars while leaping.

Inspired by dog agility competitions, we introduce Barkour (Fig.~\ref{fig:teaser}), a challenging parkour course designed specifically for quadruped robots. We propose Barkour as a comprehensive benchmark suite for evaluating the agility of legged robots. We select a few representative obstacles (the weave poles, an A-frame, a jump board, and starting/ending pause tables) and fit the setup in a \SI{25}{\meter\squared} area. We design an agility scoring system inspired by the dog competition rules: to get a high score, the legged robot must complete the entire Barkour course within a time limit determined by the configuration of the obstacles. The faster the robot, the higher the final score.

To solve the tasks in our Barkour benchmark suite, we introduce a simulation setup and two learning-based baselines as references. Our first approach involves training specialist policies in simulation that can overcome each individual obstacle. The specialist policies are then orchestrated by a high-level navigation controller that selects the appropriate specialist policy based on the location of the robot. In our second approach, we take inspiration from recent work on training generalist agents ~\cite{lee2022mgdt, reed2022generalist, brohan2022rt, baker2022vpt} and develop a transformer-based generalist locomotion policy, named \emph{Locomotion-Transformer}, which tackles all Barkour obstacles using a single policy network. To demonstrate the effectiveness of the learned agile skills, we deploy the simulation-trained policies in a zero-shot manner on a custom-built quadruped robot in a real-world Barkour setup.

Our contributions can be summarized as follows:
\begin{itemize}
    \item A benchmark (\emph{Barkour}) for agile quadruped robot locomotion inspired by dog agility competitions.
    \item Two learning-based approaches (specialist and generalist (\emph{Locomotion-Transformer}) policies) that can complete the benchmark agilely which can serve as baselines to benchmark future algorithms.
    \item A detailed analysis of zero-shot sim-to-real transfer using a custom-built quadruped robot.
\end{itemize}

\section{Related Work}
Benchmarks are a driving force behind the development of artificial intelligence methods, such as ImageNet~\cite{deng2009imagenet} for computer vision and OpenAI Gym~\cite{brockman2016openai} for Reinforcement Learning. In the field of legged robots, while many prior works focused on creating new algorithms and controllers, limited effort has been directed towards creating a systematic benchmark to assess the performance of these controllers, especially in the context of agility~\cite{torricelli2014benchmarking, eckert2019benchmarking}. Among these efforts, Eckert \& Ijspeert~\cite{eckert2019benchmarking} proposed a suite of $13$ metrics to measure different aspects of the agility of legged robots including leaping, standing, balancing, and climbing. Their metrics are carefully designed and measure the robot performance in a comprehensive way. 
However, there are a few challenges in directly leveraging these metrics to develop a general agile locomotion controller: 1) a standardized environment was not provided as part of the benchmark, making it difficult for different groups to quickly iterate on a common set of tasks, and 2) having $13$ metrics for different agility skills means that researchers may opt to focus on a subset of the metrics instead of trying to push agility as a whole. Barkour is complementary to the metrics proposed in~\cite{eckert2019benchmarking}, with the key difference being the inclusion of a standardized and extensible obstacle course environment as well as a single metric to measure the overall performance of the controller in an intuitive way. 

Combining reinforcement learning with dynamics randomization~\cite{peng2020learning, xie2021dynamics}, system ID~\cite{tan2018sim, lee2020learning}, locomotion primitives~\cite{iscen18a}, and proper reward shaping~\cite{siekmann2021sim, hwangbo2019learning}, legged robots can exhibit stable locomotion blindly on general uneven terrains at moderate speeds. 
Further advancements in perception and vision have increased the robustness and adaptability of legged robots. The pioneering work from Miki et al.~\cite{miki2022learning} demonstrated how their robot, ANYmal, can walk on various types of uneven terrain using a noisy state estimation and heightfield~\cite{miki2022learning}. Rudin et al. demonstrated that by combining these insights with a highly parallelized simulator, one can obtain a high-performing locomotion controller within minutes~\cite{rudin2022learning}. Recently, Agarwal et al.~\cite{agarwal2022legged} showed that it is possible to perform many visual locomotion tasks by consuming raw camera images in an LSTM-based network~\cite{agarwal2022legged}. Our approach builds on the findings of~\cite{miki2022learning} and leverages a GPU-based simulation environment similar to~\cite{rudin2022learning}. However, our work distinguishes itself from previous studies by emphasizing both agility and generalization, and by showcasing several agile skills performed together.

While the aforementioned works focused mainly on the robustness and generalizability of a policy, researchers have also explored methods to enable more rich and dynamic motions. Some prior works focus on enabling a certain aspect of agile locomotion, such as fast running or jumping~\cite{park2017high, cheetah3, kim2019highly, margolis2022rapid, chignoli2022rapid, nguyen2019optimized, li2022zero, iscen2021learning}. For example, Kim et al.~\citep{kim2019highly} combined Model Predictive Controller (MPC) and Whole-body Controller (WBC) to enable the MIT Mini-Cheetah quadruped robot to run at a top speed of \SI{3.7}{\meter\per\second}. Margolis et al.~\cite{margolis2022rapid} applied a learning-based approach with an automated training curriculum to further push the same robot platform to reach \SI{3.9}{\meter\per\second}. However, to solve Barkour tasks, our controllers prioritize maneuverability on complex terrains over maximizing speed on flat surfaces. In terms of handing environments with missing sections, Margolis et al.~\cite{margolis2021learning}, Yu et al.~\cite{yu2021visual}, and Lee et al.~\cite{lee2022piars} demonstrated how legged robots can conquer terrains with gaps by combining a high-level learning-based planning policy and an MPC/WBC-based motion controller, with~\cite{margolis2021learning} reporting successful jumps of a small quadruped over a \SI{0.26}{\meter} gap. Chignoli et al.~\cite{chignoli2022rapid} developed a hierarchical control framework to achieve omni-directional jumping on a quadruped robot from static pose. Li et al.~\cite{li2022zero} extended the MPC-based control pipeline to incorporate more versatile motion styles that are not limited to static poses. In contrast to these works, we demonstrate performing of a set of diverse agile skills with learning-based motion controllers, measured in a carefully designed benchmark environment. 

\section{Barkour benchmark for measuring agility}

\begin{figure}[bt]
    \centering
    \includegraphics[width=0.5\textwidth]{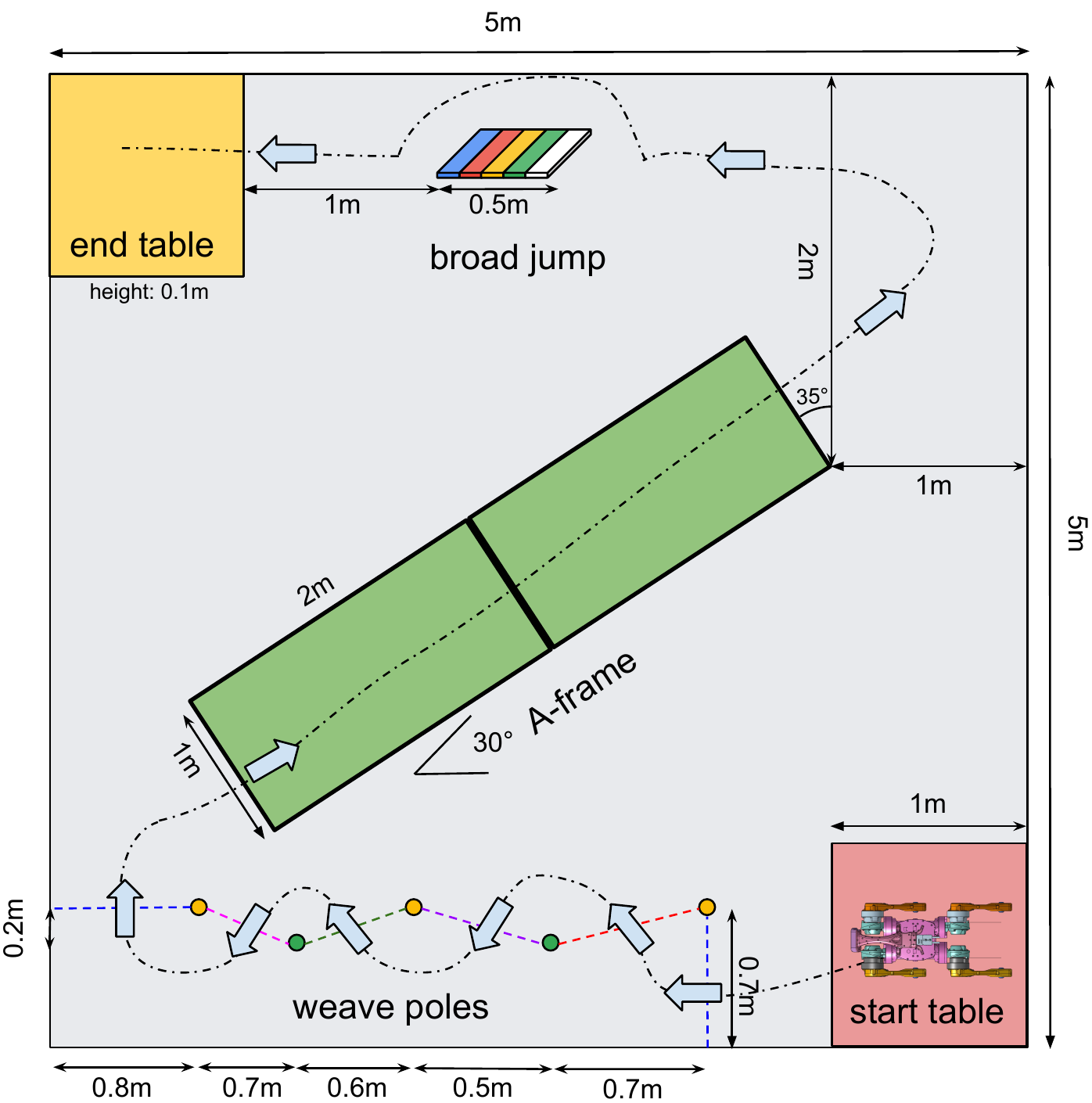}
    \caption{Barkour course design composed of four different obstacles: (start and end) pause tables, weave poles, an A-frame, and a broad jump.}
    \label{fig:barkour_spec}
\end{figure}

Making robots move like animals or humans is a goal shared by many researchers in the robotics community. Animals and their behaviors have been an inspiration for many robots. In this work we aim to make this connection quantitative by introducing  Barkour, an agility benchmark that assesses the overall agility skills of quadruped robots. The Barkour benchmark is inspired by dog agility competitions, which we believe provides a setting to quantitatively evaluate small to medium-sized quadruped robots. Dogs exhibit a wide range of agile locomotion skills and behaviors and dog agility events have well-defined rules. Dog agility competitions capture a diverse set of skills and have clear metrics based on timing and behaviors to compare performances of different dogs. A careful design of obstacles, metrics, and coverage of skill diversity is also crucial for creating a benchmark for robotics. Hence, we reuse many elements common to dog agility competitions: obstacle definitions, a time-based metric, and penalties for rules violations. 

We use an area of \SI{5}{\metre} $\times$ \SI{5}{\metre} within which we place four unique obstacles (Fig.~\ref{fig:barkour_spec}). This is a denser setup with a smaller footprint than a typical dog competition to allow for repeated deployment in a standard robotics lab. We chose the obstacle types that provide diversity over required primitive skills such as running, sideways movement, climbing and jumping, while keeping the smaller footprint.
Starting from the start table, the robot must weave through a set of poles, traverse an A-frame, jump across a broad jump and then make its way onto the end table. All obstacles are fixed in place.

\subsection{Barkour score}
The \emph{agility score} $R_\text{agility}$ measures how fast a robot can successfully complete all obstacles in Barkour. The score is calculated based on real dog competitions\footnote{\href{https://images.akc.org/pdf/rulebooks/REAGIL.pdf}{Regulations for Agility Trials and Agility Course Test (ACT) rules/scoring from the American Kennel Club (AKC)}. See standard course time for 8-inch Division Novice A and B Agility Standard Class.} with some simplifications. A score of $1.0$ indicates that the robot solved the entire course within \emph{allotted course time}  $t_\text{allotted}$. Starting from $1.0$, the robot can receive two types of deductions: a $0.1$ penalty for each failed or skipped obstacle and a $0.01$ penalty for each full second the robot exceeds  $t_\text{allotted}$. An episode is completed when the robot reaches the end table, otherwise it is terminated when $R_\text{agility}$ reaches $0$. The score\footnote{Typically represented as $[0-100]$ points in real dog competitions.} calculation is given by:
\begin{equation}
    R_\text{agility} = 1.0 - \max(t_\text{run} - t_\text{allotted}, 0) * 0.01 - \mbox{penalties}
\end{equation}

The allotted course time is the sum of the \emph{nominal size} $d_\text{obstacle}$ of each obstacle, divided by the target average speed. The nominal size is an estimate of the length of typical trajectory to complete a given obstacle, including the distance leading up to and away from it.
Based on the rule book for real dog competitions and the size of the robot, we set the target average speed to $v_\text{target}=$\SI{1.69}{\meter\per\second}. This target average speed can be scaled up for larger robots. The allotted time is calculated by:

\begin{equation}
    t_\text{allotted} = \sum_\text{obstacles} d_\text{obstacle}/v_\text{target} 
\end{equation}

In the context of this paper, we consider four types of obstacles with nominal obstacle sizes and allotted times given in Table~\ref{tab:nominal_size}. Appendix~\ref{appendix:barkour_obstacles} provides detailed definitions of each type of obstacle, including the physical setup, acceptance criteria, etc.

\begin{table}[ht]
\centering
\begin{tabular}{lll}
\textbf{Obstacle} & $d_\text{obstacle}$ & $t_\text{allotted}$ \\ \hline
pause table (x2)      & \SI{1}{\meter}  & \SI{0.59}{\second}                       \\
weave poles       & \SI{6}{\meter}  &  \SI{3.55}{\second}                    \\
A-frame           & \SI{6}{\meter}  &   \SI{3.55}{\second}               \\
broad jump        & \SI{4}{\meter}  &     \SI{2.36}{\second}              \\
\rowcolor[HTML]{EFEFEF} full course       & \SI{18}{\meter}  &    \SI{10.64}{\second}         

\end{tabular}
\caption{Nominal obstacle sizes for score calculation.\label{tab:nominal_size}}
\end{table}

Note that real dog competitions often include other types of penalties (e.g., smaller deductions if a dog retries an obstacle), whereas Barkour only deducts points for failed/skipped obstacles or excess time. This makes implementing the scoring mechanism in both sim and real much easier and significantly less error prone. 

In this work, we chose a small area with a small number of obstacles and simpler metrics for ease of repeated and controlled hardware experimentation, and kept diversity of skills and measurability as high priorities. As is the case for real dog agility competitions, our proposed metric, environment, and simulation setup can be easily adapted to a different number of obstacles, a larger area, or other variables.

\section{Two Baseline Solutions}
\label{sec:control_framework}
\begin{figure*}[bt]
    \centering
    \includegraphics[width=0.9 \textwidth]{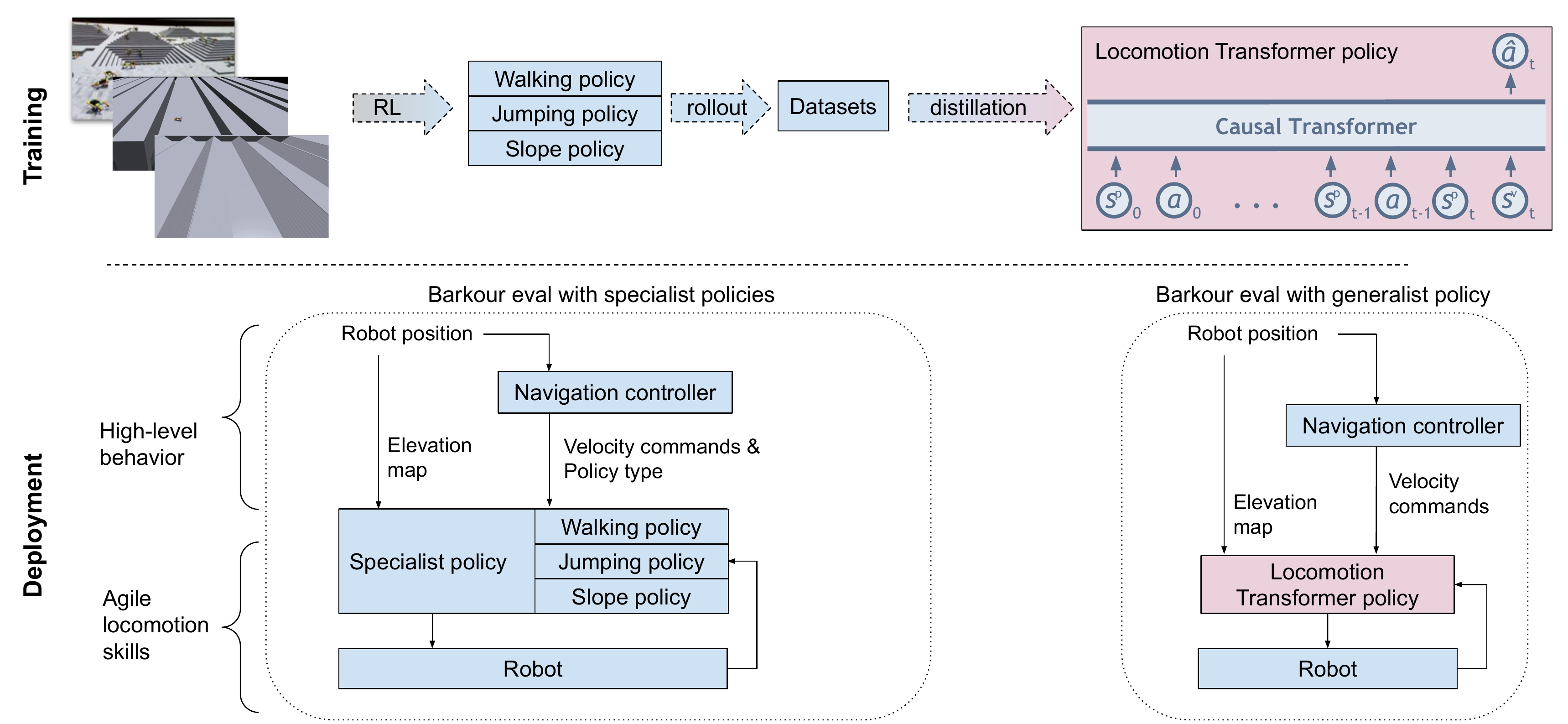}
    \caption{Overview of our methods to establish strong baselines for the Barkour benchmark. Top: Omni-directional walking, slope, and jumping policies are trained in simulation using RL. We then run the policies to create datasets which we use to distill a generalist \emph{Locomotion-Transformer} policy. Bottom left: Switching between \emph{specialist policies} using a hierarchical controller. Bottom right: \emph{generalist (Locomotion-Transformer)} trained by distilling multiple specialist policies.
    }
    \label{fig:method_overview}
\end{figure*}

One goal of this work is to demonstrate that by working towards our proposed Barkour benchmark, we can advance our techniques and insights in obtaining controllers with better agility. To achieve this, we present two learning-based methods to synthesize agile controllers for a real quadruped robot and establish baselines for the proposed Barkour benchmark. An overview of the two learning methods is shown in Fig.~\ref{fig:method_overview}.

In the first baseline, we train three specialist policies for different tasks in a physics simulation with an on-policy Reinforcement Learning algorithm (\Cref{subsec:specialist}). Each specialist policy is trained with a reward function and a terrain curriculum tailored to the corresponding task. We apply domain randomization for transferring the simulation-trained policies to the real world. We then design a high-level navigation controller (\Cref{subsec:mentor}) that switches between different specialist policies and generates velocity commands for the specialist policies to overcome different obstacles. 

In the second baseline, we train a single generalist policy that can handle all Barkour tasks. To achieve this, we collect a dataset with all the specialist policies and distill them into a unified Transformer-based locomotion policy, which we name \textit{Locomotion-Transformer}. We then combine Locomotion-Transformer with a simplified navigation controller that only needs to provide the waypoints for the low-level policy to tackle the full obstacle course.

\subsection{Specialist Policy Learning}
\label{subsec:specialist}

Our first baseline trains three specialist policies to cover all the core agility skills required in the Barkour benchmark: fast omni-directional walking on uneven terrain, climbing up and down a slope, and jumping over a board. The first policy is used to solve the start/end tables and weave pole tasks, the second solves the A-Frame task, and the third solves the Broad Jumping task. All the specialist policies are trained using PPO~\cite{schulman2017proximal} in LeggedGym~\cite{rudin2022learning}, which uses the GPU-powered NVidia IsaacGym. Specialist model architecture details can be found in Appendix~\ref{app:reward_function}. We also investigated the TPU-powered Brax simulator ~\cite{freeman2021brax} for training, as presented in Appendix~\ref{appendix:brax}.

\subsubsection{Observation and Action Space}

The observation space of the specialist policies include desired base velocities $\bar{v} = (\bar{v_x}, \bar{v_y}, \bar{\omega_z})$, robot proprioceptive information $s^\text{p}=(\mathbf{g}, \theta, \omega_z)$, terrain perception $s^\text{v}$, and the last robot action $a_{t-1}$, where $\bar{v_x}, \bar{v_y}, \bar{\omega_z}$ are the desired linear velocity (front/back, left/right) and the desired angular velocity in yaw in the robot's local frame, $\mathbf{g}$ is the projected gravity vector in the robot local frame, $\theta$ is the robot joint angles, and $\omega_z$ is the yaw velocity. Given that we are working with a partially-observable system (POMDP), we further include a observation history of length \SI{0.3}{\second} for the robot proprioception.

In our approach, perception is modeled as a heightfield in the vicinity of the robot following the method used by Miki et al~\cite{miki2022learning}. To construct the heightfield $s^\text{v}$, we select a grid of locations on the terrain surrounding the robot and measure their relative heights with respect to the center of the robot's torso. This grid translates and rotates with the robot, always maintaining alignment with its heading. The shape of the heightfield is customized to suit the requirements of each task, which are elaborated in the corresponding sections below.

We choose the action space to be the target motor angles measured from a nominal standing pose.

\subsubsection{Reward Function}
Similar to Rudin et al.~\cite{rudin2022learning}, our reward function consists of multiple terms that fall into two categories: Task and Regularization. Task rewards encourage the robot to perform the desired skills, such as running forward or turning, while regularization rewards shape the behavior of the robot, such as low energy consumption and high stability. ~\Cref{app:reward_function} lists the reward terms that we use in this work.

\subsubsection{Omni-directional Walking Policy}
The Omni-directional Walking Policy (OWP) is trained to follow a velocity command in all directions on uneven terrains. It demonstrates the ability of the controller to quickly adjust the speed profile and the robot's orientation, which are critical skills for the weave pole and pause table tasks.

OWP takes all three categories of observations as input and is trained with randomly sampled velocity commands. The training environment for OWP follows the general uneven terrain curriculum designed by Rudin et al \cite{rudin2022learning}, which consists of mild slopes, stairs, and random steps. Please refer to Appendix~\ref{app:owp_observation} for details of the observation space, the velocity sampling, and the reward structure.

\subsubsection{Slope Climbing Policy}

For the A-frame task, the robot must agilely climb up and down a \SI{30}{\degree} slope. Due to the large angle of the slope, the robot operates near the limit of available traction. It needs to precisely manage the trade-off between its speed and its ability to maintain friction with the ground. We find that simply including the target slope terrain during OWP training cannot solve this task.

To tackle this task, we train a specialized Slope Climbing Policy (SCP). SCP shares the same observation space, actions space, and reward function as OWP. However, it differs from OWP training in two aspects. First, SCP is trained with a curriculum of increasing slope angles from \SI{5}{\degree} to \SI{33}{\degree}, as shown in the leftmost simulation image in Fig.~\ref{fig:method_overview}. Second, since the key skill required in the A-Frame task is to climb up and down the slope, we can focus the training on moving agilely in the frontal direction instead of training the policy to move fast in all directions. For details on the terrain curriculum and the training config, see Appendix~\ref{app:scp_training}. 

\subsubsection{Jumping Policy}
In the jumping task, the robot needs to jump over a board that is \SI{0.5}{\meter} long, which is longer than its body. This requires not only moving fast, but also stepping precisely (as close to the broad jump as possible, but not on it). To train the Jumping Policy (JP), we use a 3-stage curriculum, including flat terrain running, gap training, and fine-tuning with more randomizations (Appendix~\ref{app:jp_training}). 

\begin{table}[t]
\scriptsize
	\begin{center}
		\begin{tabular}{ll}
			\toprule
                \textbf{Parameters} & \textbf{Randomization Range} \\
                \midrule
                Torso mass & $[2.0, 6.5]$\SI{}{\kilo\gram}\\
                Torso inertia & $[40\%, 165\%]$\\
                Torso linear velocity perturbation & $[0, 1]$ m/s updated at \SI{10}{\second} intervals\\
                Ground friction & $[0.5, 1.25]$\\
                Position gain & $[15, 20]$\SI{}{\newton\meter\per\radian}\\
                Damping gain & $[0.5, 0.75]$ \SI{}{\newton\meter\second\per\radian}\\
                Joint static friction & $[0, 0.7]$\\
			\bottomrule
		\end{tabular}
	\end{center}
	\captionsetup{font={small,it},labelsep=colon}
	\caption{Domain Randomization parameters. \label{tab:randomization_parameters}}
\end{table}

\subsubsection{Domain Randomization}
To bridge the sim-to-real gap, we employ the technique of domain randomization \cite{peng2018sim}. The randomized parameters are listed in Table \ref{tab:randomization_parameters}. We find that the default domain randomization scheme designed by Rudin et al.~\cite{rudin2022learning} works well when the robot velocity is relatively low (slower than \SI{1}{\meter\per\second}). However, for more agile motions, such as jumping or slope climbing, where the speed of the robot can exceed \SI{2}{\meter\per\second}, we observe a notable sim-to-real gap.
To overcome this challenge, we incorporate additional randomization including torso inertia, motor modeling, and joint static friction. Torso inertia randomization enables the agent to better control the orientation of the body at high speed, and motor modeling and joint static friction are known important factors for quadruped sim-to-real \cite{tan2018sim}. We find these to be critical for successful transfer of policies to the real hardware.

\subsection{Locomotion-Transformer: A Generalist Locomotion Policy}
\label{subsec:generalist}

Our second baseline aims to train a single generalist locomotion policy that can tackle all obstacles, thereby removing the necessity for ad-hoc switching between individual specialist policies and promoting generalization capabilities to different obstacle and terrain configurations. Our generalist policy, \emph{Locomotion-Transformer}, is trained by distilling the specialist policies via behavioral cloning with a Transformer sequence model~\cite{vaswani2017attention}, similar to~\citep{lee2022mgdt,brohan2022rt,reed2022generalist}.
Although prior work~\cite{espeholt2018impala, rudin2022advanced} has demonstrated that generalist policies can also be learned via reinforcement learning by simultaneously training on multiple environments, the diverse curricula and reward structures in our problem setting makes it challenging.

\subsubsection{Data Collection}
Collecting the right interaction data that covers the state distribution on which the policy is going to operate is critical to successful policy distillation~\cite{brohan2022rt}.
To learn the generalist, we collect an offline dataset by rolling out individual specialist policies in simulated environments that are same as the training settings of the specialist. 
We collect $17,636$ simulated episodes, equivalent to $57.58$ hours of robot time, across four different types of environments: random steps, stairs, gaps, and slopes. 
More details on the data collection setup used in this work and a data card (\Cref{tab:transfurmer_data_card}) can be found in Appendix~\ref{app:data_collection}.

\subsubsection{Transformer Model}
As shown in \Cref{fig:method_overview}, Locomotion-Transformer is a causal Transformer that takes a history of velocity commands $\bar{v}$, proprioceptive states $s^\text{p}_{0:t}$, actions $a_{0:t-1}$, and the most recent heightfields $s^\text{v}_{t}$ over a fixed context window as inputs in the following order: $$\{ s^\text{p}_{0}, a_{0}, \ldots, s^\text{p}_{t-1}, a_{t-1}, s^\text{p}_{t}, s^\text{v}_t \}$$
$s^\text{v}$ contains all types of heightfields that are originally designed for each individual environment (see details in \Cref{subsec:specialist}).
The model predicts the next action $a_t$ at the last position, and is trained on an L2 regression loss.
We tokenize the most recent elevation image with a two-layer convolutional encoder network for each type of heightfield.
The proprioceptive states (along with the velocity commands) and actions are each tokenized with one projection layer.
We use a context window of $0.3$s, the same as in  the specialist policy, which amounts to a size of $W=15$. For terrain perception, we combine both heightfields from OWP and JP to cover the terrain near and in front of the robot. 
We use ReLU activation to encode each observation input.
The model architecture hyperparameters can be found in Appendix~\ref{app:transfurmer_architecture}. 

\subsection{High Level Navigation Controller}
\label{subsec:mentor}
\begin{figure}[bt]
    \centering
    \includegraphics[width=0.4\textwidth]{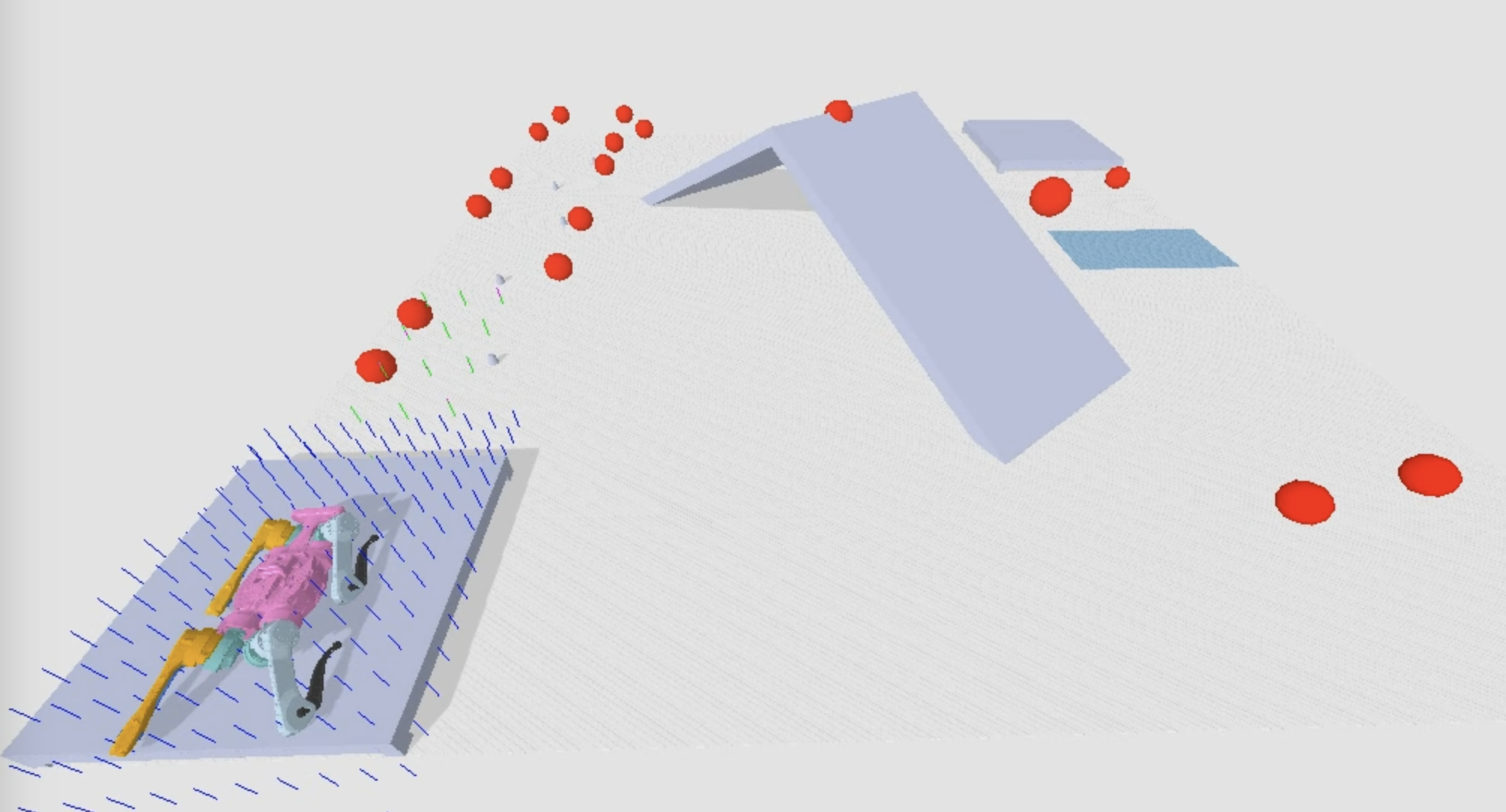}
    \caption{An example set of way points used for the navigation controller. The robot stands on the start table and its heightfield is illustrated with blue rays hitting the floor.}
    \label{fig:mentor}
\end{figure}

The specialist policies allow the quadruped to tackle different individual obstacles. However, to complete the Barkour benchmark, the robot must also make decisions on how to navigate the field and transition between behaviors while following specific rules of the obstacle course.
To achieve this, we design a high-level state machine-based navigation controller to command our locomotion policies and guide the robot through the entire course. The navigation controller has access to the full state of the environment and the ground truth robot pose, and outputs appropriate velocity commands for the robot.

Specifically, we place a sequence of waypoints around the obstacles that serve as sub-goals to guide the robot through each obstacle in the Barkour benchmark. Each waypoint specifies a desired position and heading orientation for the robot to reach. Every timestep, the high-level navigation controller computes linear and angular velocity commands to take the robot to the desired poses. An error tolerance is also encoded in each waypoint to determine when to switch to the next waypoint. When using the specialist policies, each waypoint also includes a behavior type to dictate which specialist policy to use. On the other hand, the generalist Locomotion-Transformer policy only needs the velocity commands. More details about this calculation can be found in Appendix~\ref{appendix:Navigation Controller}.

\section{Robot Hardware}
Exploring the influence of the Barkour benchmark on enhancing the agility of quadruped robots necessitates considerable controller development and thorough real hardware experimentation. This poses significant challenges on the reliability and repeatability of the robot hardware, especially given the highly agile movements we strive for. Moreover, quadruped animals exhibit diverse body configurations compared to typical quadruped robots, which can significantly affect their capacity for agile motion. Consequently, we believe that hardware optimization and customization are critical in bridging the agility gap between legged robots and their animal counterparts. 

\begin{figure}[ht]
    \centering
    \includegraphics[scale=0.075]{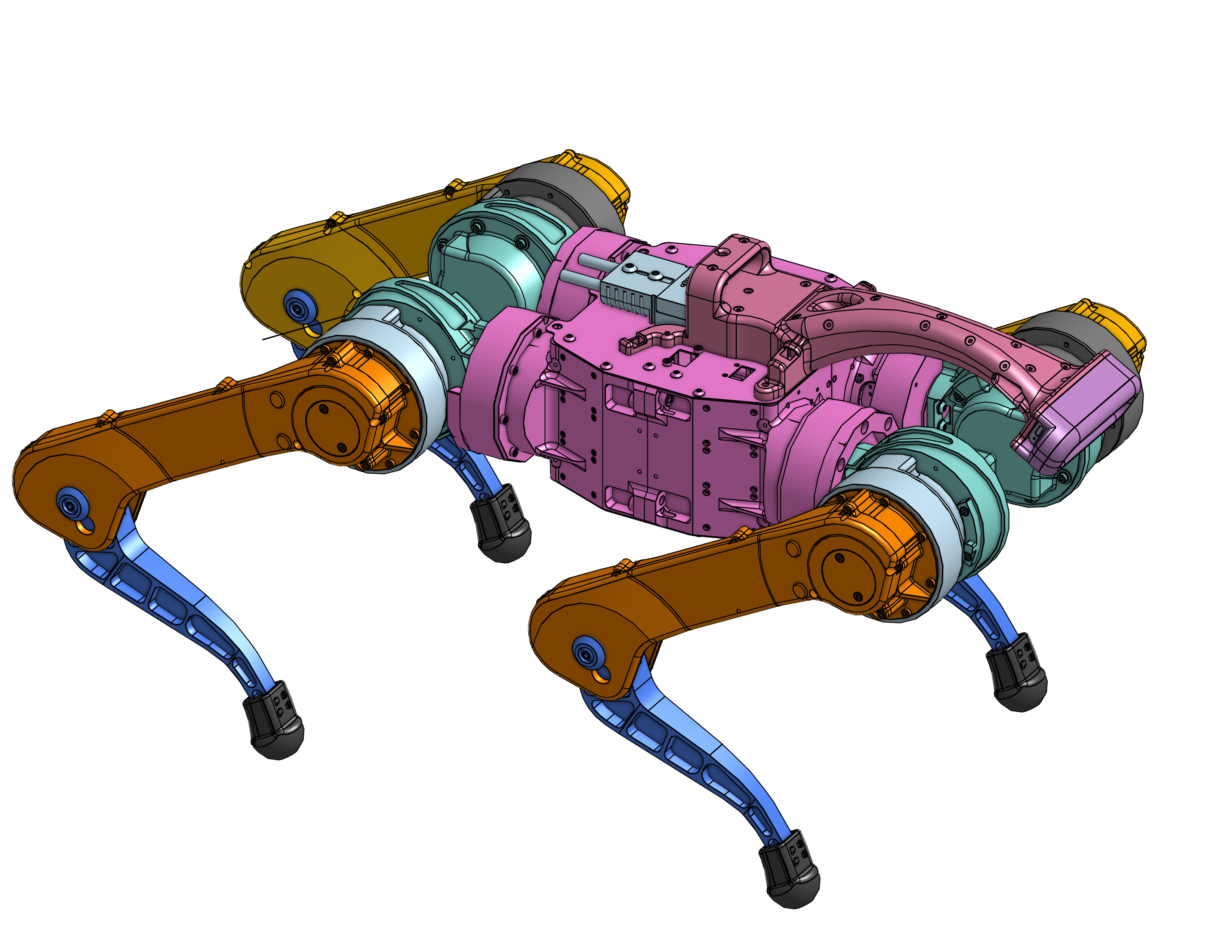}
    \caption{Custom small quadruped robot for hardware evaluation.}
    \label{fig:robot_cad}
\end{figure}

As a result, we developed a small quadruped robot in-house to evaluate its learned agile skills using the Barkour benchmark.
The robot (Fig.~\ref{fig:robot_cad}) is similar in size to the Unitree A1 and MIT Mini-Cheetah robots.
The robot weighs \SI{11.5}{\kilo\gram}, and has \SI{220}{\milli\meter} upper limbs and \SI{190}{\milli\meter} lower limbs. The front-to-back hip-to-hip distance is \SI{380}{\milli\meter} and the distance between the left and right legs is \SI{290}{\milli\meter}.

We use \href{https://store.tmotor.com/goods.php?id=981}{T-Motor AK80-6} actuators, which are controlled by Elmo G-SOLTWIR50/100SE2S motor drivers running at \SI{24}{\volt}. The actuators provide a peak output torque of \SI{12}{\newton\meter} at each joint.

For our experiments, we use both the robot's onboard sensors (Parker 3DMCX5-AHRS IMU, joint position, velocity, torque) as well as an external motion capture system (\href{https://www.phasespace.com/x2e-motion-capture/}{Phasespace X2E}) to track the robot's position and orientation in the obstacle course. 
We control the robot using an off-board workstation with two Intel Xeon Gold 6154 CPUs via a CAN bus connected to each leg. 
The learned policies send position commands at \SI{50}{\hertz}, and the PD control loop runs at \SI{1}{\kilo\hertz}. For all hardware experiments, the PD gains are set to (\SI{20}{\newton\meter\per\radian}, \SI{0.5}{\newton\meter\second\per\radian}).

As part of the development of the benchmark and methods, we evaluated approximately 3600 Barkour attempts across two robots, which corresponds to about 24 hours of continuous operation time and approximately \SI{60}{\kilo\meter} of robot moving distance.

\section{Experiments and Discussions}
We evaluate the specialist and generalist policies within our hierarchical control framework on Barkour tasks. We aim to answer the following questions:
\begin{itemize}
 \item \emph{Is Barkour a good benchmark for agility?}
 \item \emph{Can we solve Barkour benchmark using the frameworks proposed in Section \ref{sec:control_framework} and how does it compare to animal agility?}
 \item \emph{How important are the design choices we made during specialist and generalist policy training?}
\end{itemize}

\subsection{Why Barkour as a Benchmark}

The benchmark is effective for several reasons. Firstly, it requires diverse motion capabilities to complete the course, which effectively exposes potential limitations on agile skill discovery. Secondly, the benchmark effectively tests the maneuverability and control precision of the locomotion controller at high speeds due to the complex routes required to finish the course as illustrated in Fig.~\ref{fig:eval_trajectory}, and scoring being tied to the completion time of the course. In the event that the robot misses the correct gate in the weave poles section, touches the jump board, or if the robot is too slow, the overall score is penalized. Lastly, we found that the benchmark is well-suited for real animal counterparts, as demonstrated in our experiments involving two small dogs. The comparison between the performance of the real dogs and robots highlights opportunities for improvement in both hardware (such as the need for flexible spines) and algorithmic approaches. It is worth noting that the real dogs never failed to complete any of the obstacles and were significantly faster than our provided baselines.

\subsection{Barkour Evaluation using Combined Specialist Policies}

\begin{figure}[ht]
    \centering
    \includegraphics[width=0.3\textwidth]{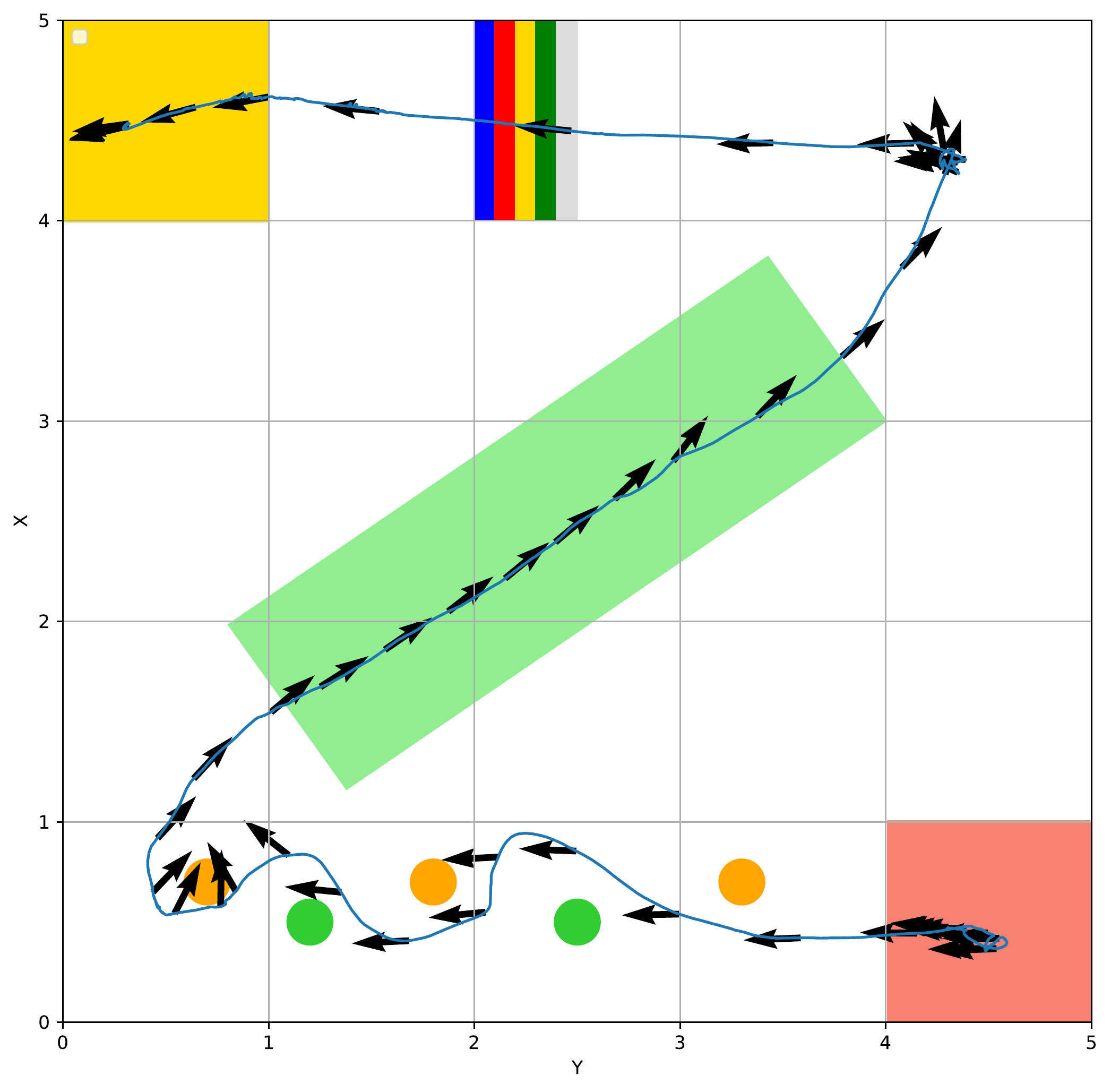}
    \caption{Top-down view of the robot's trajectory during a single high performance evaluation where the robot completes the course in \SI{20.4}{\second} scoring 0.91. The arrows show the robot's heading at \SI{0.5}{\second} intervals.}
    \label{fig:eval_trajectory}
\end{figure}

\begin{figure}[t]
    \centering
    \includegraphics[width=0.5\textwidth]{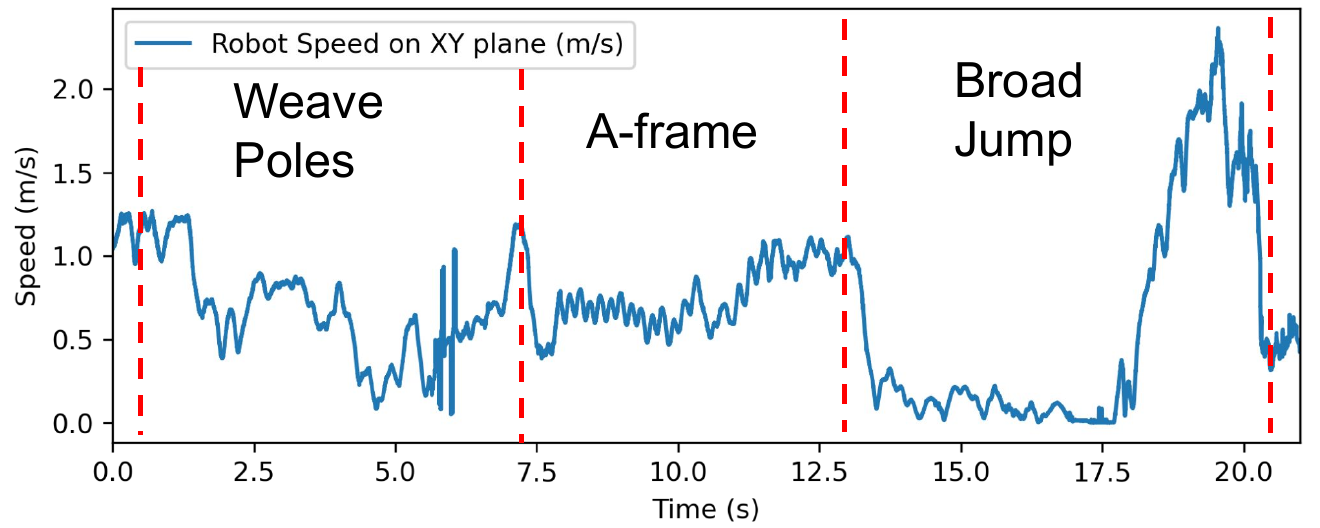}
    \caption{Analysis of the same run as Fig.~\ref{fig:eval_trajectory} showing the robot's speed.}
    \label{fig:eval_analysis}
\end{figure}

First, we evaluate the performance of specialist policies combined on all Barkour tasks. We show a sample behavior of the robot in Fig.~\ref{fig:eval_trajectory}. In this run, the robot completes the course by traversing all of the obstacles in \SI{20.4}{\second}. While the robot completes all the obstacles, this specific run receives 0.91 agility score due to the extra time it took compared to the allotted time of \SI{10.64}{\second}. One can see that the robot completes the weave poles section at a \SI{90}{\degree} angle, which is enforced by our high level controller, due to the narrow passages between the last few poles. The velocity of the robot during the same run is shown in Fig.~\ref{fig:eval_analysis}. The robot's forward velocity varies between \SI{0.5}{\meter\per\second} and \SI{1}{\meter\per\second} during the weave poles. It reaches a velocity as high as \SI{2.3}{\meter\per\second} during the jumping phase. For comparison, we ran the same course with two small dogs: a Pomeranian/Chihuahua (\SI{3.2}{\kilo\gram}, \SI{33}{\centi\meter} height at withers) and a Dachshund (\SI{4}{\kilo\gram}, \SI{20}{\centi\meter} height at withers). After getting familiar with the course, they both reached a maximum score ($1.0$) by reaching the final pause table in approximately \SI{9}{\second} (videos in supplementary).

\begin{table}[ht]
\scriptsize
	\begin{center}
		\begin{tabular}{llll}
			\toprule
                \textbf{Policy Type} & \textbf{Barkour Score} & \textbf{Completion Time} & \textbf{Forward Velocity} \\
                \midrule
                Specialist & $0.77 \pm 0.064$ & $24.6\pm 1.1$ s & $0.74 \pm 0.04$ m/s \\
                Locomotion-Transformer & $0.73 \pm 0.062$ & $25.8\pm 1.6$ s & $0.69 \pm 0.04$ m/s \\
                Small dog & $1.00 \pm 0$ & $9.02 \pm 0.65$ s & \\
			\bottomrule
		\end{tabular}
	\end{center}
	\captionsetup{font={small,it},labelsep=colon}
	\caption{Results for full Barkour course with specialist policies and the generalist Locomotion-Transformer. We report mean and one std dev for each metric.}
	\label{tab:barkour_results}
\end{table}

\begin{table}[ht]
\scriptsize
	\begin{center}
		\begin{tabular}{llll}
			\toprule
                &  \textbf{Weave Poles} & \textbf{A-Frame} & \textbf{Broad Jump}\\
                \midrule
                Completion Time & $9.27 \pm 0.87$\SI{}{\second} & $7.95 \pm 0.73$\SI{}{\second} & $2.45 \pm 0.54$\SI{}{\second} \\
                Nominal Size & \SI{6}{\meter} & \SI{6}{\meter} & \SI{4}{\meter} \\ 
                Forward Velocity & $0.73 \pm 0.06$ \SI{}{\meter\per\second} & $0.68 \pm 0.06$ \SI{}{\meter\per\second} & $1.7 \pm 0.24$  \\
                Success Rate & $100\%$ & $100\%$ & $38\%$ \\ 
			\bottomrule
		\end{tabular}
	\end{center}
	\captionsetup{font={small,it},labelsep=colon}
	\caption{Results for specialist policies with individual obstacles. We report mean and one std dev for each metric. \label{tab:specialist_results}}
\end{table}

\begin{figure}
    \centering
    \includegraphics[width=0.48\textwidth]{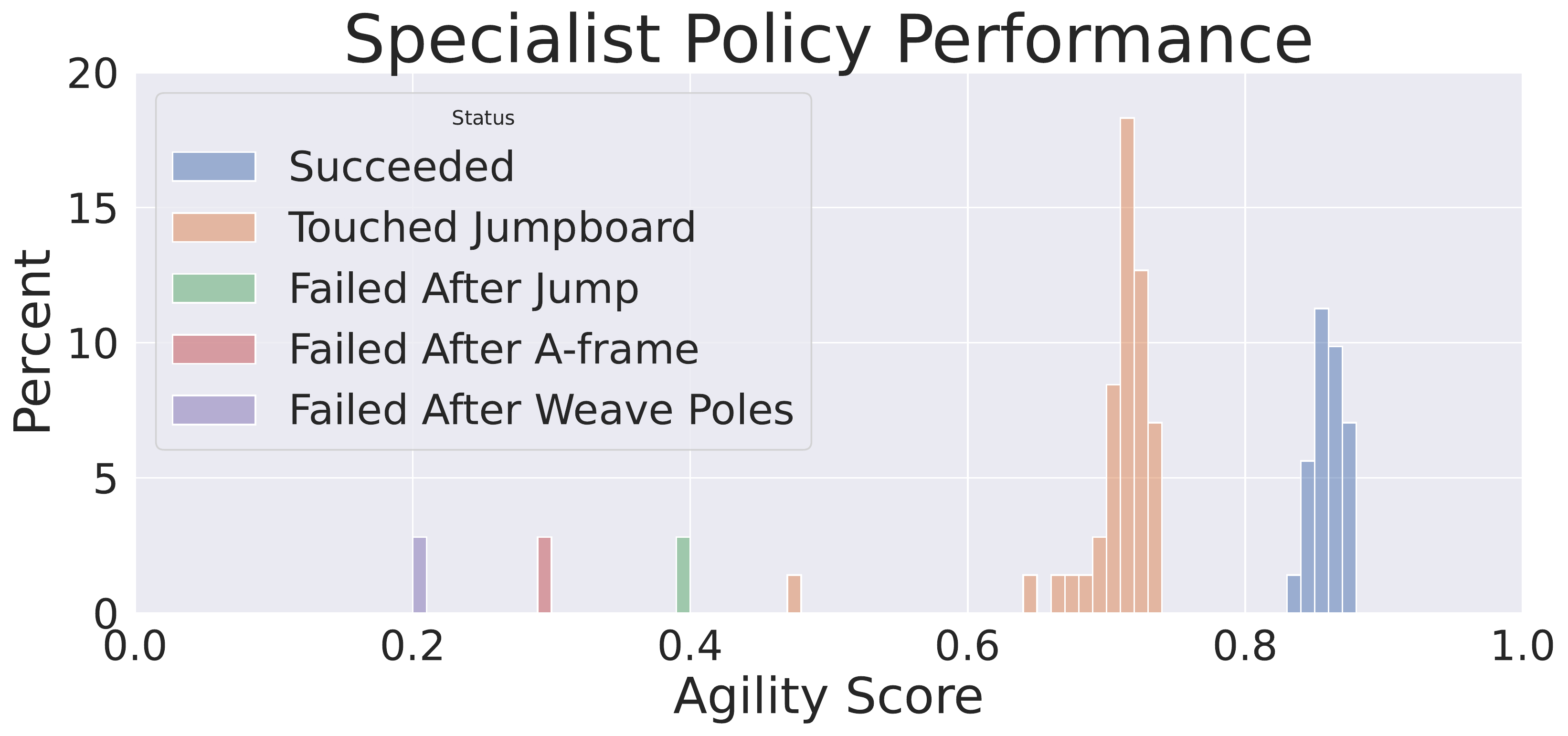}
    \includegraphics[width=0.48\textwidth]{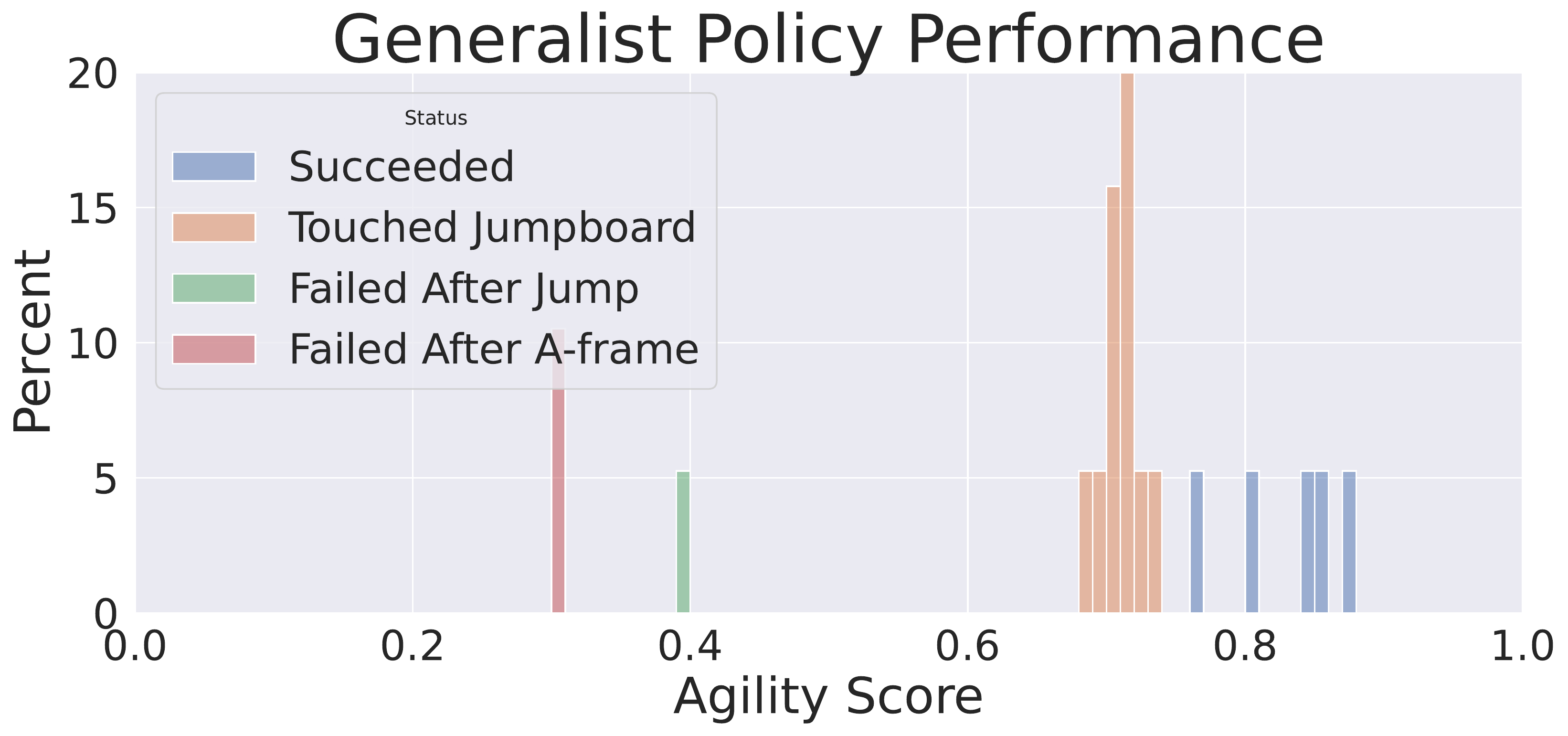}
    \caption{Statistical results for Barkour episodes with specialist policies. Top: Distribution of the scores for 70 trials using specialist policies. Bottom: Distribution of the scores for 19 trials using generalist transformer-locomotion architecture.}
    \label{fig:eval_statistical}
\end{figure}

We test the robustness of the hierarchical controller by running it $71$ times. Fig.~\ref{fig:eval_statistical} shows that the robot completes the course in most of these runs, while falling down or tipping over 6 times. In $40$ of the runs, the robot touches the jump board and still scores approximately $0.73$. In $25$ of these runs, the robot completes all 5 obstacles scoring $0.87$ on average. As shown in Table~\ref{tab:barkour_results}, the mean score of all the runs is around $0.77$ with an average base forward velocity of \SI{0.74}{\meter\per\second}.

The decomposition of these runs (Table~\ref{tab:specialist_results} into the different obstacles shows that the specialist policies complete the weave poles at around the nominal (target) speed, while the broad jump is faster and the A-frame is slower. The robot completes the weave poles and A-frame with a 100\% success rate, while the broad jump, requiring particularly more agile behavior, succeeds 38\% of the time. We also provide the detailed analysis of each specialist policy below:

\subsubsection{Omni-directional Walking Policy}
This policy is capable of producing consistent trotting motions in both simulation and real hardware. The motion is highly stable when the robot is commanded to move between \SI{-1.5}{\meter\per\second} and \SI{1.5}{\meter\per\second} in the forward and backward directions and between \SI{-1}{\meter\per\second} and \SI{1}{\meter\per\second} laterally with a yaw rate between \SI{-2.5}{\radian\per\second} and \SI{2.5}{\radian\per\second} on flat ground. With the policy, the robot can step up and down steps of up to \SI{0.1}{\meter},
which corresponds to about one third of the robot's standing height. During our $60$ specialist policy trials, OWP succeeds at the weave poles every time, taking \SI{9.27}{\second} on average. Compared with real dogs that can complete this section in less than \SI{4}{\second}, OWP takes significantly more time because (1) the robot does not have a flexible spine, and (2) the robot is much wider and has to re-orient itself sideways to fit the narrow passages between the poles.  

\subsubsection{Slope Climbing Policy}
This policy can run on the steep \SI{30}{\degree} A-frame using a trotting gait while maintaining a speed of more than \SI{0.6}{\meter\per\second}. The same policy climbs down the A-frame reaching \SI{1.5}{\meter\per\second}. The policy exhibits consistent performance ($100\%$ success rate) regardless of variations in the initial position or angle of attack as long as the robot remains within the width of the A-frame.

\subsubsection{Jumping Policy}
\begin{figure}
    \centering
    \includegraphics[width=0.48\textwidth]{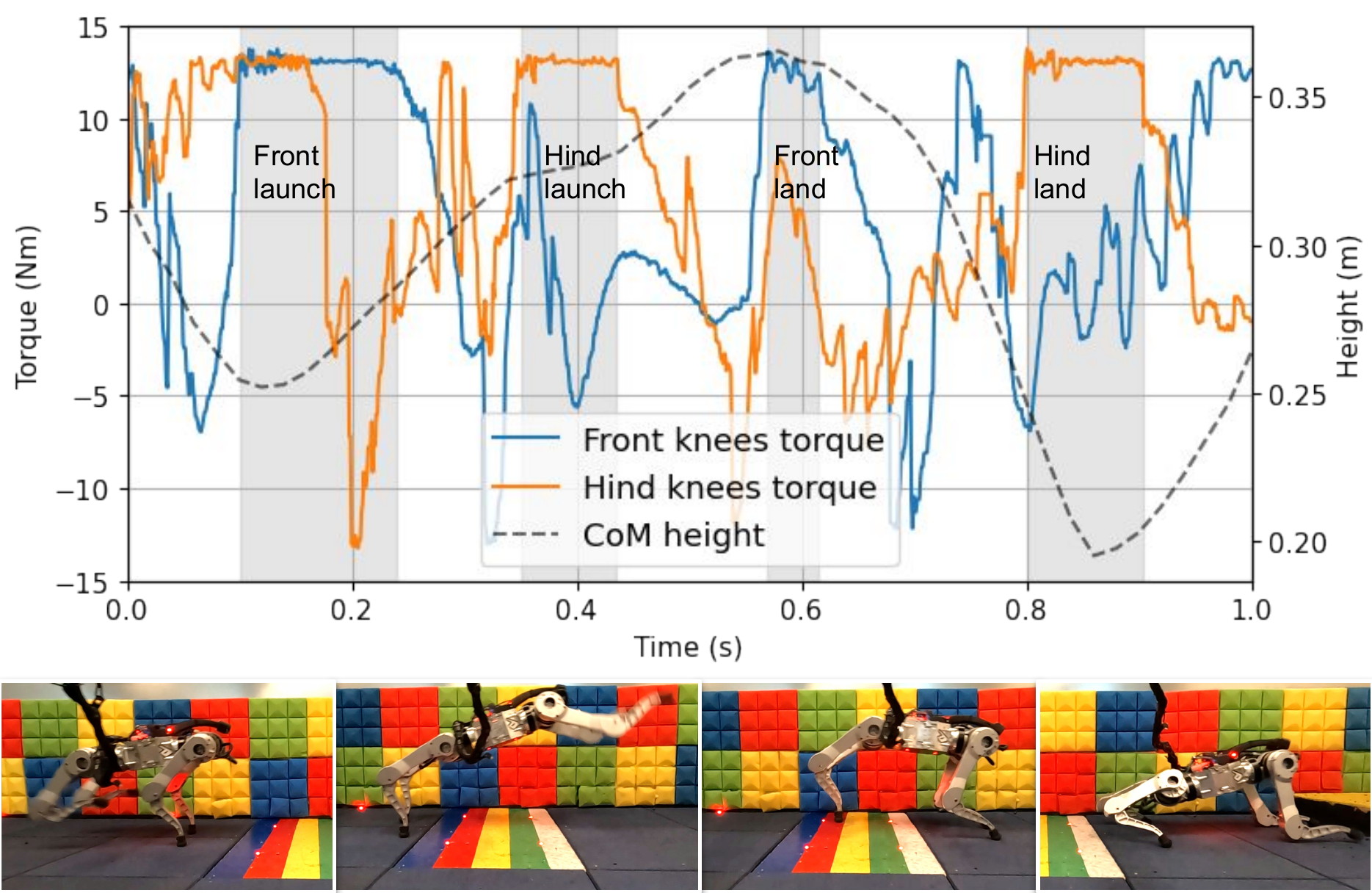}
    \caption{Broad jump example. Top: Average front and hind knee torque during various phases of the jump. Bottom: Main jump phases: accelerating, lift-off, landing the front legs, landing the rear legs.}
    \label{fig:jump-torque}
\end{figure}

This policy demonstrates a bounding gait when traversing flat terrain. To clear the broad jump gap of \SI{0.5}{\meter}, the robot accelerates from nearly stationary at a distance of \SI{2}{\meter} in front of the jump and reaches a  velocity of \SI{2.3}{\meter\per\second} at the jump. An example of the center of mass speed and trajectory can be observed in Fig.~\ref{fig:eval_analysis}. In 38\% of trials, the robot is able to clear the \SI{0.5}{\meter} gap, using its momentum to jump farther than would be possible with a standing jump. In most of the failure cases, the robot touches the edge of the jump board either before lift-off or after landing. A typical example of the policy on the broad jump obstacle is shown in Fig.~\ref{fig:jump-torque}. 

\subsection{Barkour Evaluation using Locomotion-Transformer}
We now evaluate the performance of the generalist policy, which is distilled using the data generated by the specialist policies. 
Unlike the combined specialist policies that require the high-level navigation policy to select which specialist policy to use, our Locomotion-Transformer model absorbs all the low-level skills of the specialists and thus can be deployed on the robot without explicit hints on which obstacle it is handling. Therefore, we use a simplified navigation policy that only outputs the velocity command to the policy during evaluation.

We evaluate Locomotion-Transformer on the real Barkour setup with $19$ trials, which can be seen in \Cref{fig:eval_statistical}. In general, we observe slightly lower performance for the Locomotion-Transformer policy compared to the combined specialist policies (also seen in \Cref{tab:barkour_results}). A major reason for this is that the Locomotion-Transformer policy does not have information about the obstacle type being tackled and has to infer relevant information from terrain and proprioceptive observations, which increases the difficulties of the task. 

On the other hand, we want to highlight two advantages of the generalist Locomotion-Transformer policy. First, by using a single generalist policy, we achieve smoother transitions between different obstacles. As seen in the supplementary video, when transitioning between different specialist policies, the robot may exhibit jerky motions when the previous specialist policy reaches a state that is unfamiliar to the following policy. Meanwhile, the Locomotion-Transformer achieves smooth transitions among different behaviors including different gaits.

\begin{figure}
    \centering
    \includegraphics[width=0.29\textwidth]{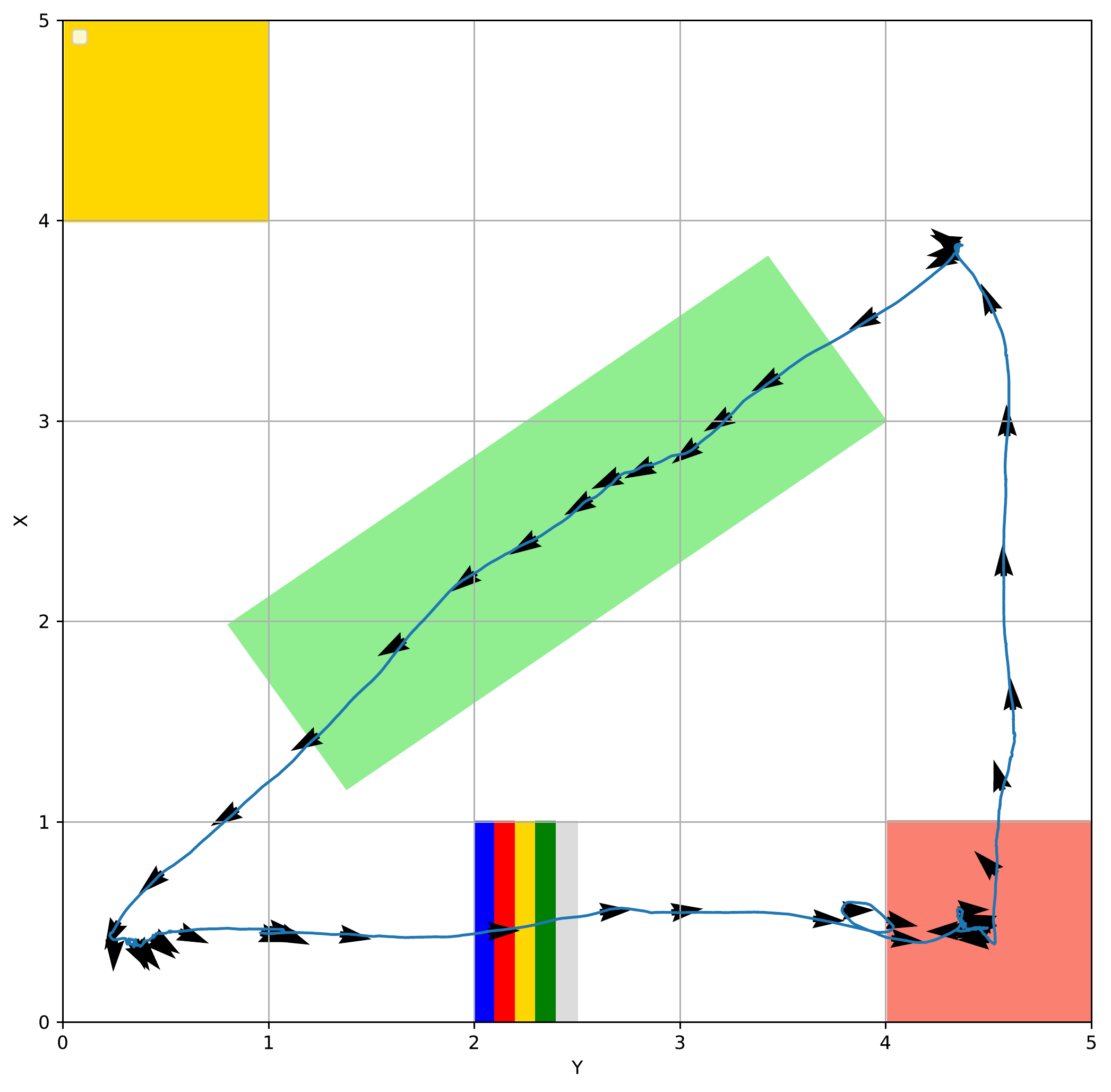}
    \caption{Generalization aspect of the Locomotion-Transformer policy tested on a modified course. We change the waypoints and obstacle locations. While the policy is not aware of the obstacle types, the robot can still climb up the A-frame from the opposite end or achieve the broad jump in different locations.}
    \label{fig:generalization_test}
\end{figure}

Another advantage of the Locomotion-Transformer policy is that it's more general and flexible. By removing the reliance on explicitly knowing the obstacle type, the Locomotion-Transformer works for a wider range of environments without requiring domain expertise to specify which policy to use. To demonstrate the flexibility and generalization capability of Locomotion-Transformer, we adjusted the order of the Barkour obstacles to create a new course: the robot must perform fast \SI{90}{\degree} turning, then go to the end of the A-Frame and perform the A-Frame task backwards. Next, the robot must do a broad jump towards the start table and climb up the table. With specialist policies, one would need to re-assign different policies to different sub-goals and fine-tune the transitions between different policies; the Locomotion-Transformer can generalize to this new route by simply providing a set of new sub-goals to denote the desired route.
The results in Fig.~\ref{fig:generalization_test} show that the robot succeeds in climbing the A-frame from the opposite end and going over the broad jump in a different location.

We have shown that the generalist Locomotion-Transformer policy is able to execute a wide range of high-level commands provided by a navigation controller and achieves a high Barkour score. However, the training is based on simulation data from specific expert policies, which raises two follow-up questions:
\begin{itemize}
    \item \emph{Is the Transformer-based framework also capable of learning high-level behaviors}?
    \item \emph{Is it possible to train a Locomotion-Transformer policy from a hardware dataset}? 
\end{itemize}

We answer these questions affirmatively in Appendix~\ref{appendix:hwdistill} by training and deploying a course-specific Locomotion-Transformer policy from hardware data.

\subsection{Is Training Specialist Policies Necessary?}
We have demonstrated that it is possible to train individual specialist policies for each task and distill them into one generalist policy to solve the Barkour benchmark. However, one question remains: \emph{Is it possible to train a single agent using RL that combines the capabilities of all specialist policies \autoref{subsec:specialist}?}

To answer this question, we train a single multi-task policy using our IsaacGym-based training pipeline. We construct a training terrain curriculum by mixing the terrains from all three specialist policy types. During RL training, $50\%$ of the data are from the slope environment for training Slope Climbing Policy, $30\%$ are from the general uneven terrain for training Omni-directional Walking Policy, and $20\%$ are from the gap environment for the Jumping Policy. We follow the same curriculum for each terrain type as described in \Cref{subsec:specialist}. We use the reward function from training the Omni-directional Walking Policy to obtain a policy that takes a velocity command as input.

We deploy the trained policy on the real robot and find that the resulting policy can robustly walk down the start table and finish the weaving pole task. However, it cannot climb up the A-Frame even though we have already adjusted the training data distribution to bias towards the slope task. This demonstrates that the steep slope necessitates the use of a specialist training. In addition, the policy cannot successfully perform the broad jump. This experiment also demonstrates the difficulties and diversity of the skills required to solve the proposed Barkour benchmark.

\subsection{Locomotion-Transformer Ablations}
\label{app:model_ablations}

\begin{figure}[t!]
    \centering
    
    \begin{subfigure}{0.34\columnwidth}
        \centering
        \includegraphics[width=\linewidth]{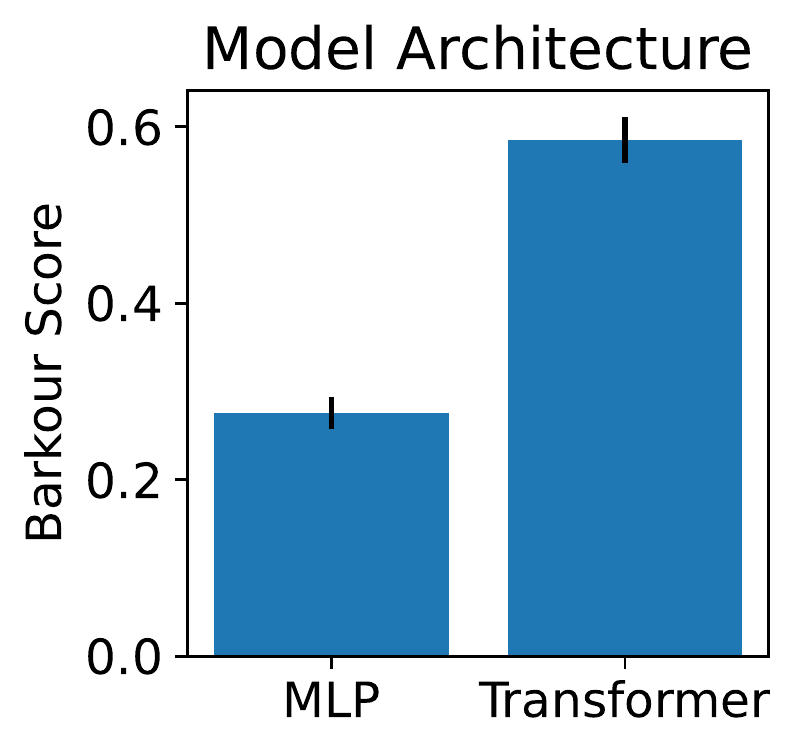}
        \caption{Model architecture.}
        \label{sub:model_architecture}
    \end{subfigure}  
    ~
    \begin{subfigure}{0.6\columnwidth}
        \centering
        \includegraphics[width=\linewidth]{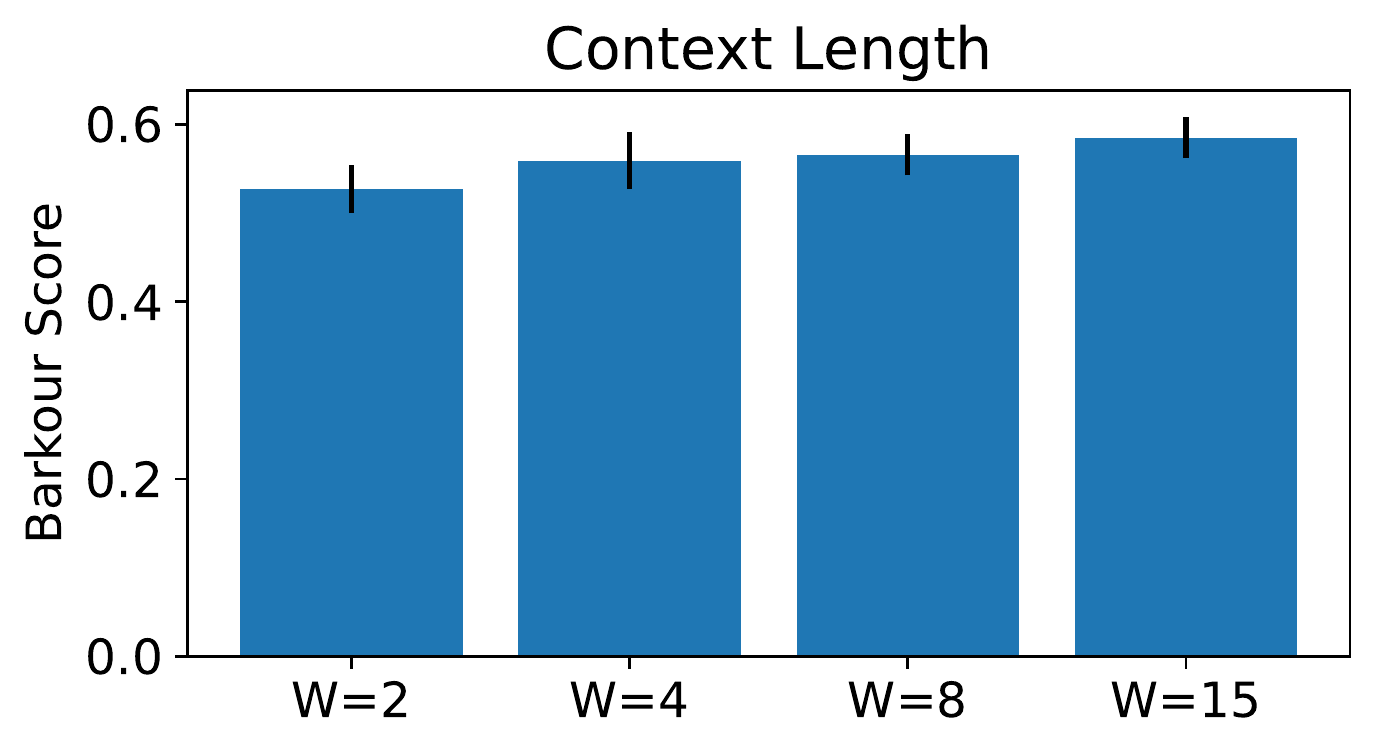}
        \caption{Context length.}
        \label{sub:context_length}
    \end{subfigure}    
    
    \begin{subfigure}{0.47\columnwidth}
        \centering
        \includegraphics[width=\linewidth]{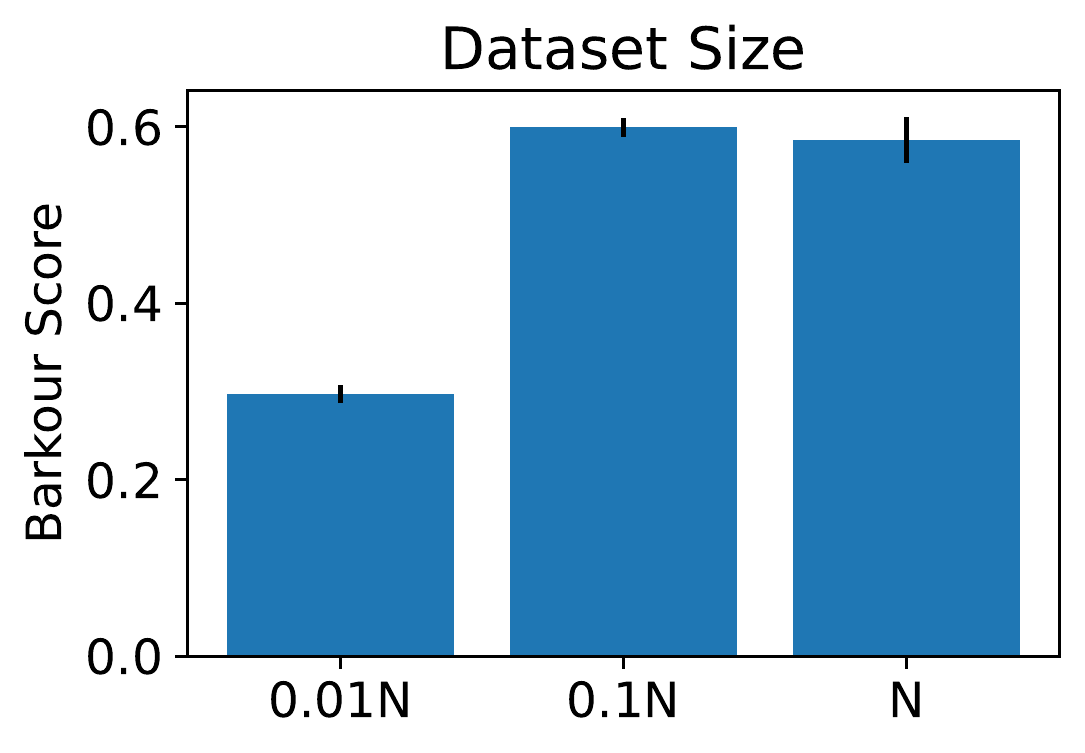}
        \caption{Dataset size.}
        \label{sub:dataset_size}
    \end{subfigure}        
    ~
    \begin{subfigure}{0.47\columnwidth}
        \centering
        \includegraphics[width=\linewidth]{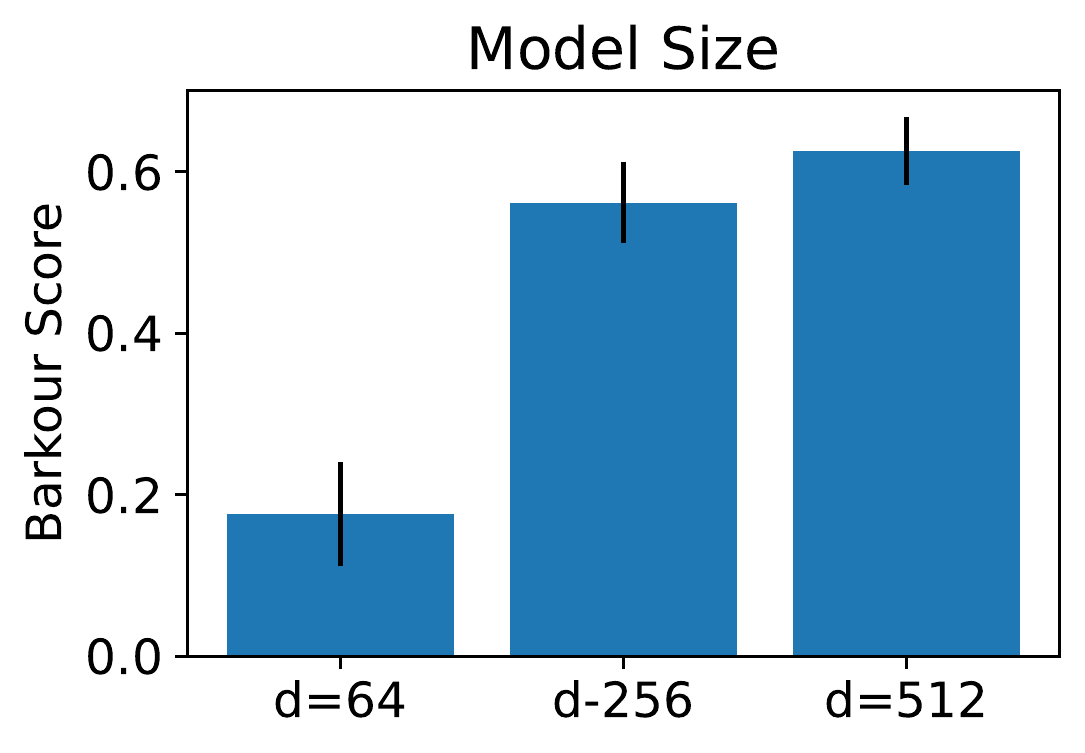}
        \caption{Model size.}
        \label{sub:model_size}
    \end{subfigure}    
    
    \caption{Ablation results (evaluated in simulation).}
    \label{fig:ablations}
\end{figure}

We investigate several design choices behind Locomotion-Transformer, including model architecture, model size, training dataset size, and context length. To enable rapid model iteration, we report PyBullet simulation Barkour scores. For each experiment, we train three models with different random seeds and report the mean and standard deviation of the respective mean evaluation scores.

\subsubsection{Model architecture} To understand if the transformer architecture is necessary, we trained an MLP (a larger version of the specialist policy architecture) on the same dataset. As shown in Fig.~\ref{sub:model_architecture}, the distilled MLP policy performs substantially worse than the Transformer policy. 

\subsubsection{Context length} In Fig.~\ref{sub:context_length}, we show the Locomotion-Transformer performance with different context lengths. Using a longer context length is more effective, suggesting that the Transformer is able to leverage past state information.

\subsubsection{Dataset size} In Fig.~\ref{sub:dataset_size}, we train Locomotion-Transformer models with 1\%, 10\%, and 100\% of the training data. While 1\% ($\sim$176 episodes) is insufficient for training a competitive policy, 10\% ($\sim$1760 episodes) and 100\% result in similar performance. We note that using a larger or more diverse dataset may lead to improved sim-to-real performance, which we leave as investigation for future work.

\subsubsection{Model size} Fig.~\ref{sub:model_size} shows that increasing the size of the Locomotion-Transformer model can lead to improved performance, which mirrors similar results in other domains such as language. However, deploying these models on real robots places strict bounds on inference latency and therefore model size.

\section{Conclusion}
We present Barkour, a benchmark to evaluate the agility of quadruped robots. Inspired by dog agility competitions, Barkour is a testbed with an intuitive scoring mechanism that requires combination of various agile skills. Furthermore, it can be easily adapted to robots of various sizes or extended by adding or rearranging obstacles while retaining the same metrics.

To set a strong baseline and make progress towards the Barkour benchmark, we explore a learning-based sim-to-real approach and proposed two baseline solutions. In both solutions, we first train a set of specialist policies that excel at tackling individual tasks by leveraging recent developments in fast simulation techniques with on-policy RL algorithms. While in the first solution, we manually design a state machine to switch between different specialist policies. In the second solution,
we distill these skills into a generalist Transformer-based policy, named Locomotion-Transformer, that can automatically and smoothly transition between different obstacles. Our results show that Locomotion-Transformer can exhibit multiple agile skills required by the proposed benchmark and can automatically switch between them depending on the sensed environment and the commands received from a high-level navigation controller. As a validation of our approach, we also demonstrate that the policies can generalize to other obstacle configurations.

We believe that providing a benchmark for legged robotics, especially for agility, is an important first step to quantify the progress towards animal-level agility for quadruped robots. The Barkour benchmark is far from solved. While our baseline solutions can reach a peak agility score of 0.91 and complete the course in approx. \SI{20}{\second}, an untrained dog achieved 1.0 agility score and can complete the course in about half the robot's time (approx. \SI{9}{\second}). There is still a big gap in agility between robots and their animal counterparts, as demonstrated in this benchmark. Additionally, the fact that state-of-the-art RL methods fail to learn a single policy to complete the course further underscored the complexity and value of Barkour as a benchmark. We believe that Barkour will serve the robotics community as an important testbed for different learning and control methods and different hardware designs.

\section{Limitations and Future Work}
Our proposed approach sets a strong baseline on the benchmark. However, as the scores reflect, Barkour is not fully solved and there is still notable room to push towards dog-level agility by improving speed and robustness. We believe that bridging this gap necessitates a collective endeavor from the research community, and the suggested Barkour benchmark can help effectively track this progress.

One limitation of the current proposed baseline methods is that we use privileged information such as the CAD model of the environment and the position of the robot (via a Motion-Capture system) in the world frame. An important future work direction is to explore Barkour using only on-board sensors for both low-level locomotion skills and high-level navigation controller.

An equally exciting direction for future research on Barkour is to evaluate the impact of modifications to robot hardware, different form-factors, and sensors on performance or training speed. Finally, we are also looking into evaluating Barkour in an interactive setting, closer to real-world dog agility competitions, with a human leading a robot through the course. 

\section*{Acknowledgments}
We would like to thank Marissa	Giustina,
Gus	Kouretas,
nubby	Lee,
James	Lubin,
Sherry	Moore,
Thinh	Nguyen,
Krista	Reymann,
Satoshi	Kataoka,
Trish	Blazina and the rest of the
robotics team at Google DeepMind for their feedback and contributions.

\bibliographystyle{plain}
\bibliography{references}

\begin{thebibliography}{10}

\bibitem{agarwal2022legged}
Ananye Agarwal, Ashish Kumar, Jitendra Malik, and Deepak Pathak.
\newblock Legged locomotion in challenging terrains using egocentric vision.
\newblock {\em arXiv preprint arXiv:2211.07638}, 2022.

\bibitem{baker2022vpt}
Bowen Baker, Ilge Akkaya, Peter Zhokov, Joost Huizinga, Jie Tang, Adrien
  Ecoffet, Brandon Houghton, Raul Sampedro, and Jeff Clune.
\newblock {Video PreTraining (VPT)}: Learning to act by watching unlabeled
  online videos.
\newblock In {\em Advances in Neural Information Processing Systems}, 2022.

\bibitem{cheetah3}
Gerardo Bledt, Matthew~J. Powell, Benjamin Katz, Jared Di~Carlo, Patrick~M.
  Wensing, and Sangbae Kim.
\newblock Mit cheetah 3: Design and control of a robust, dynamic quadruped
  robot.
\newblock In {\em 2018 IEEE/RSJ International Conference on Intelligent Robots
  and Systems (IROS)}, pages 2245--2252, 2018.

\bibitem{brockman2016openai}
Greg Brockman, Vicki Cheung, Ludwig Pettersson, Jonas Schneider, John Schulman,
  Jie Tang, and Wojciech Zaremba.
\newblock Openai gym, 2016.

\bibitem{brohan2022rt}
Anthony Brohan, Noah Brown, Justice Carbajal, Yevgen Chebotar, Joseph Dabis,
  Chelsea Finn, Keerthana Gopalakrishnan, Karol Hausman, Alex Herzog, Jasmine
  Hsu, et~al.
\newblock Rt-1: Robotics transformer for real-world control at scale.
\newblock {\em arXiv preprint arXiv:2212.06817}, 2022.

\bibitem{chignoli2022rapid}
Matthew Chignoli, Savva Morozov, and Sangbae Kim.
\newblock Rapid and reliable quadruped motion planning with omnidirectional
  jumping.
\newblock In {\em 2022 International Conference on Robotics and Automation
  (ICRA)}, pages 6621--6627. IEEE, 2022.

\bibitem{choi2023learning}
Suyoung Choi, Gwanghyeon Ji, Jeongsoo Park, Hyeongjun Kim, Juhyeok Mun,
  Jeong~Hyun Lee, and Jemin Hwangbo.
\newblock Learning quadrupedal locomotion on deformable terrain.
\newblock {\em Science Robotics}, 8(74):eade2256, 2023.

\bibitem{akc2023}
American~Kennel Club.
\newblock {\em Regulations for Agility Trials and Agility Course Test (ACT)}.
\newblock American Kennel Club, 2023.

\bibitem{deng2009imagenet}
Jia Deng, Wei Dong, Richard Socher, Li-Jia Li, Kai Li, and Li~Fei-Fei.
\newblock Imagenet: A large-scale hierarchical image database.
\newblock In {\em 2009 IEEE conference on computer vision and pattern
  recognition}, pages 248--255. Ieee, 2009.

\bibitem{dicarlo2018}
Jared Di~Carlo, Patrick~M. Wensing, Benjamin Katz, Gerardo Bledt, and Sangbae
  Kim.
\newblock Dynamic locomotion in the mit cheetah 3 through convex
  model-predictive control.
\newblock In {\em 2018 IEEE/RSJ International Conference on Intelligent Robots
  and Systems (IROS)}, pages 1--9, 2018.

\bibitem{eckert2019benchmarking}
Peter Eckert and Auke~J Ijspeert.
\newblock Benchmarking agility for multilegged terrestrial robots.
\newblock {\em IEEE Transactions on Robotics}, 35(2):529--535, 2019.

\bibitem{espeholt2018impala}
Lasse Espeholt, Hubert Soyer, Remi Munos, Karen Simonyan, Vlad Mnih, Tom Ward,
  Yotam Doron, Vlad Firoiu, Tim Harley, Iain Dunning, et~al.
\newblock Impala: Scalable distributed deep-rl with importance weighted
  actor-learner architectures.
\newblock In {\em International conference on machine learning}, pages
  1407--1416. PMLR, 2018.

\bibitem{freeman2021brax}
C~Daniel Freeman, Erik Frey, Anton Raichuk, Sertan Girgin, Igor Mordatch, and
  Olivier Bachem.
\newblock Brax--a differentiable physics engine for large scale rigid body
  simulation.
\newblock {\em arXiv preprint arXiv:2106.13281}, 2021.

\bibitem{hutter2016anymal}
Marco Hutter, Christian Gehring, Dominic Jud, Andreas Lauber, C~Dario
  Bellicoso, Vassilios Tsounis, Jemin Hwangbo, Karen Bodie, Peter Fankhauser,
  Michael Bloesch, et~al.
\newblock Anymal-a highly mobile and dynamic quadrupedal robot.
\newblock In {\em 2016 IEEE/RSJ international conference on intelligent robots
  and systems (IROS)}, pages 38--44. IEEE, 2016.

\bibitem{hwangbo2019learning}
Jemin Hwangbo, Joonho Lee, Alexey Dosovitskiy, Dario Bellicoso, Vassilios
  Tsounis, Vladlen Koltun, and Marco Hutter.
\newblock Learning agile and dynamic motor skills for legged robots.
\newblock {\em Science Robotics}, 4(26):eaau5872, 2019.

\bibitem{iscen18a}
Atil Iscen, Ken Caluwaerts, Jie Tan, Tingnan Zhang, Erwin Coumans, Vikas
  Sindhwani, and Vincent Vanhoucke.
\newblock Policies modulating trajectory generators.
\newblock In {\em Proceedings of The 2nd Conference on Robot Learning},
  volume~87 of {\em Proceedings of Machine Learning Research}, pages 916--926.
  PMLR, 29--31 Oct 2018.

\bibitem{iscen2021learning}
Atil Iscen, George Yu, Alejandro Escontrela, Deepali Jain, Jie Tan, and Ken
  Caluwaerts.
\newblock Learning agile locomotion skills with a mentor.
\newblock In {\em 2021 IEEE International Conference on Robotics and Automation
  (ICRA)}, pages 2019--2025. IEEE, 2021.

\bibitem{katz2019mini}
Benjamin Katz, Jared Di~Carlo, and Sangbae Kim.
\newblock Mini cheetah: A platform for pushing the limits of dynamic quadruped
  control.
\newblock In {\em 2019 international conference on robotics and automation
  (ICRA)}, pages 6295--6301. IEEE, 2019.

\bibitem{kim2019highly}
Donghyun Kim, Jared Di~Carlo, Benjamin Katz, Gerardo Bledt, and Sangbae Kim.
\newblock Highly dynamic quadruped locomotion via whole-body impulse control
  and model predictive control.
\newblock {\em arXiv preprint arXiv:1909.06586}, 2019.

\bibitem{kumar2021rma}
Ashish Kumar, Zipeng Fu, Deepak Pathak, and Jitendra Malik.
\newblock Rma: Rapid motor adaptation for legged robots.
\newblock {\em arXiv preprint arXiv:2107.04034}, 2021.

\bibitem{lee2020learning}
Joonho Lee, Jemin Hwangbo, Lorenz Wellhausen, Vladlen Koltun, and Marco Hutter.
\newblock Learning quadrupedal locomotion over challenging terrain.
\newblock {\em Science robotics}, 5(47):eabc5986, 2020.

\bibitem{lee2022mgdt}
Kuang-Huei Lee, Ofir Nachum, Sherry Yang, Lisa Lee, C~Daniel Freeman, Sergio
  Guadarrama, Ian Fischer, Winnie Xu, Eric Jang, Henryk Michalewski, et~al.
\newblock Multi-game decision transformers.
\newblock In {\em Advances in Neural Information Processing Systems}, 2022.

\bibitem{lee2022piars}
Kuang-Huei Lee, Ofir Nachum, Tingnan Zhang, Sergio Guadarrama, Jie Tan, and
  Wenhao Yu.
\newblock {PI-ARS: Accelerating Evolution-Learned Visual-Locomotion with
  Predictive Information Representations}.
\newblock In {\em 2022 IEEE/RSJ International Conference on Intelligent Robots
  and Systems (IROS)}, pages 1447--1454. IEEE, 2022.

\bibitem{li2022zero}
He~Li, Wenhao Yu, Tingnan Zhang, and Patrick~M Wensing.
\newblock Zero-shot retargeting of learned quadruped locomotion policies using
  hybrid kinodynamic model predictive control.
\newblock In {\em 2022 IEEE/RSJ International Conference on Intelligent Robots
  and Systems (IROS)}, pages 11971--11977. IEEE, 2022.

\bibitem{margolis2021learning}
Gabriel~B Margolis, Tao Chen, Kartik Paigwar, Xiang Fu, Donghyun Kim, Sangbae
  Kim, and Pulkit Agrawal.
\newblock Learning to jump from pixels.
\newblock {\em arXiv preprint arXiv:2110.15344}, 2021.

\bibitem{margolis2022rapid}
Gabriel~B Margolis, Ge~Yang, Kartik Paigwar, Tao Chen, and Pulkit Agrawal.
\newblock Rapid locomotion via reinforcement learning.
\newblock {\em arXiv preprint arXiv:2205.02824}, 2022.

\bibitem{miki2022learning}
Takahiro Miki, Joonho Lee, Jemin Hwangbo, Lorenz Wellhausen, Vladlen Koltun,
  and Marco Hutter.
\newblock Learning robust perceptive locomotion for quadrupedal robots in the
  wild.
\newblock {\em Science Robotics}, 7(62):eabk2822, 2022.

\bibitem{nguyen2019optimized}
Quan Nguyen, Matthew~J Powell, Benjamin Katz, Jared Di~Carlo, and Sangbae Kim.
\newblock Optimized jumping on the mit cheetah 3 robot.
\newblock In {\em 2019 International Conference on Robotics and Automation
  (ICRA)}, pages 7448--7454. IEEE, 2019.

\bibitem{park2017high}
Hae-Won Park, Patrick~M Wensing, and Sangbae Kim.
\newblock High-speed bounding with the mit cheetah 2: Control design and
  experiments.
\newblock {\em The International Journal of Robotics Research}, 36(2):167--192,
  2017.

\bibitem{peng2018sim}
Xue~Bin Peng, Marcin Andrychowicz, Wojciech Zaremba, and Pieter Abbeel.
\newblock Sim-to-real transfer of robotic control with dynamics randomization.
\newblock In {\em 2018 IEEE international conference on robotics and automation
  (ICRA)}, pages 3803--3810. IEEE, 2018.

\bibitem{peng2020learning}
Xue~Bin Peng, Erwin Coumans, Tingnan Zhang, Tsang-Wei Lee, Jie Tan, and Sergey
  Levine.
\newblock Learning agile robotic locomotion skills by imitating animals.
\newblock {\em arXiv preprint arXiv:2004.00784}, 2020.

\bibitem{radford2019language}
Alec Radford, Jeffrey Wu, Rewon Child, David Luan, Dario Amodei, Ilya
  Sutskever, et~al.
\newblock Language models are unsupervised multitask learners.
\newblock {\em OpenAI blog}, 1(8):9, 2019.

\bibitem{reed2022generalist}
Scott Reed, Konrad Zolna, Emilio Parisotto, Sergio~Gomez Colmenarejo, Alexander
  Novikov, Gabriel Barth-Maron, Mai Gimenez, Yury Sulsky, Jackie Kay,
  Jost~Tobias Springenberg, et~al.
\newblock A generalist agent.
\newblock {\em arXiv preprint arXiv:2205.06175}, 2022.

\bibitem{rudin2022advanced}
Nikita Rudin, David Hoeller, Marko Bjelonic, and Marco Hutter.
\newblock Advanced skills by learning locomotion and local navigation
  end-to-end.
\newblock In {\em 2022 IEEE/RSJ International Conference on Intelligent Robots
  and Systems (IROS)}, pages 2497--2503, 2022.

\bibitem{rudin2022learning}
Nikita Rudin, David Hoeller, Philipp Reist, and Marco Hutter.
\newblock Learning to walk in minutes using massively parallel deep
  reinforcement learning.
\newblock In {\em Conference on Robot Learning}, pages 91--100. PMLR, 2022.

\bibitem{schulman2017proximal}
John Schulman, Filip Wolski, Prafulla Dhariwal, Alec Radford, and Oleg Klimov.
\newblock Proximal policy optimization algorithms.
\newblock {\em arXiv preprint arXiv:1707.06347}, 2017.

\bibitem{siekmann2021sim}
Jonah Siekmann, Yesh Godse, Alan Fern, and Jonathan Hurst.
\newblock Sim-to-real learning of all common bipedal gaits via periodic reward
  composition.
\newblock In {\em 2021 IEEE International Conference on Robotics and Automation
  (ICRA)}, pages 7309--7315. IEEE, 2021.

\bibitem{smith2022legged}
Laura Smith, J~Chase Kew, Xue~Bin Peng, Sehoon Ha, Jie Tan, and Sergey Levine.
\newblock Legged robots that keep on learning: Fine-tuning locomotion policies
  in the real world.
\newblock In {\em 2022 International Conference on Robotics and Automation
  (ICRA)}, pages 1593--1599. IEEE, 2022.

\bibitem{tan2018sim}
Jie Tan, Tingnan Zhang, Erwin Coumans, Atil Iscen, Yunfei Bai, Danijar Hafner,
  Steven Bohez, and Vincent Vanhoucke.
\newblock Sim-to-real: Learning agile locomotion for quadruped robots.
\newblock {\em arXiv preprint arXiv:1804.10332}, 2018.

\bibitem{tassa2018deepmind}
Yuval Tassa, Yotam Doron, Alistair Muldal, Tom Erez, Yazhe Li, Diego de~Las
  Casas, David Budden, Abbas Abdolmaleki, Josh Merel, Andrew Lefrancq, et~al.
\newblock Deepmind control suite.
\newblock {\em arXiv preprint arXiv:1801.00690}, 2018.

\bibitem{torricelli2014benchmarking}
Diego Torricelli, Rahman~SM Mizanoor, Jose Gonzalez, Vittorio Lippi, Georg
  Hettich, Lorenz Asslaender, Maarten Weckx, Bram Vanderborght, Strahinja
  Dosen, Massimo Sartori, et~al.
\newblock Benchmarking human-like posture and locomotion of humanoid robots: a
  preliminary scheme.
\newblock In {\em Biomimetic and Biohybrid Systems: Third International
  Conference, Living Machines 2014, Milan, Italy, July 30--August 1, 2014.
  Proceedings 3}, pages 320--331. Springer, 2014.

\bibitem{vaswani2017attention}
Ashish Vaswani, Noam Shazeer, Niki Parmar, Jakob Uszkoreit, Llion Jones,
  Aidan~N Gomez, {\L}ukasz Kaiser, and Illia Polosukhin.
\newblock Attention is all you need.
\newblock {\em Advances in neural information processing systems}, 30, 2017.

\bibitem{xie2021dynamics}
Zhaoming Xie, Xingye Da, Michiel Van~de Panne, Buck Babich, and Animesh Garg.
\newblock Dynamics randomization revisited: A case study for quadrupedal
  locomotion.
\newblock In {\em 2021 IEEE International Conference on Robotics and Automation
  (ICRA)}, pages 4955--4961. IEEE, 2021.

\bibitem{yu2021visual}
Wenhao Yu, Deepali Jain, Alejandro Escontrela, Atil Iscen, Peng Xu, Erwin
  Coumans, Sehoon Ha, Jie Tan, and Tingnan Zhang.
\newblock Visual-locomotion: Learning to walk on complex terrains with vision.
\newblock In {\em 5th Annual Conference on Robot Learning}, 2021.

\end{thebibliography}

\clearpage

\appendix
\subsection{Author Contributions}
\begin{itemize}
\item \textbf{Methods (conception, benchmark definition, architecture development, implementation, policy training, ablations...)}
Ken Caluwaerts, Atil Iscen, J. Chase Kew, Kuang-Huei Lee, Lisa Lee, Wenhao Yu, Tingnan Zhang, Vincent Zhuang.
\item \textbf{Software infrastructure}
Ken Caluwaerts, Erwin Coumans, Daniel Freeman, Atil Iscen, J. Chase Kew, Yuheng Kuang, Lisa Lee, Ofir Nachum, Ken Oslund, Francesco Romano,  Wenhao Yu, Tingnan Zhang, Daniel Zheng.
\item \textbf{Hardware development}
Ken Caluwaerts, Jason Powell, Stefano Saliceti, Jeff Seto, Ron Sloat, Daniel Zheng.
\item \textbf{Leadership (managed or advised on the project)}
Ken Caluwaerts, Adil Dostmohamed, Raia Hadsell, Nicolas Heess, Atil Iscen, Bauyrjan Jyenis, Michael Neunert, Francesco Nori, Carolina Parada, Stefano Saliceti, Jie Tan,  Vikas~Sindhwani, Vincent Vanhoucke, Daniel Zheng.
\item \textbf{Paper (writing, figures, visualizations)}
Ken Caluwaerts, Daniel Freeman, Nicolas Heess, Atil Iscen, J. Chase Kew, Kuang-Huei Lee, Lisa Lee, Stefano~Saliceti, Jie Tan, Wenhao Yu, Tingnan Zhang,  Vincent Zhuang.
\item \textbf{Operations (data collection, hardware maintenance)}
Omar Cortes, Linda Luu, Jason Powell, Diego Reyes, Ron~Sloat.
\end{itemize}
\subsection{Barkour Definitions}
\label{appendix:barkour_obstacles}

\subsubsection{Pause Tables}
The pause table is a \SI{1}{\meter} x \SI{1}{\meter} x \SI{0.1}{\meter} (width, depth, height) solid obstacle. The top consists of a hard (Shore B 70-90), non-slip surface. There is one pause table at the start of the course (start table) and one at the end (end table).

To successfully step off a table, the robot must move the center of its torso  \SI{0.7}{\meter} away from the center of the table. To successfully step onto the a table, the robot must move the center of its torso to within  \SI{0.4}{\meter} of the center of the table.

\subsubsection{Weave Poles}
The weave poles obstacle consists of a series of 5 small flexible poles spaced between \SI{0.5}{\meter} and \SI{0.8}{\meter} apart. The fixed part of each cone is approximately \SI{0.05}{\meter} in diameter and approximately \SI{0.05}{\meter} high. In order the achieve the obstacle, the robot must go between the poles while maintaining a distance of \SI{0.1}{\meter} between its center and the poles.

\subsubsection{A-frame}
The A-frame consists of a ramp with a \SI{30}{\degree} incline. The top of the obstacle is \SI{1}{\meter} high and each side is \SI{1}{\meter} wide. 
We use artificial turf as surface for the A-frame, which increases diversity of surfaces while providing enough traction to climb.
The gap between the floor and the flat part of the ascending and descending ramps shall be less than \SI{20}{\milli\meter}. 

To successfully complete the A-frame obstacle, the robot must pass (center of torso) across the line segment defined by the part of the A-frame touching floor on the first side, then move across the top of the A-frame and finally across the line segment at the bottom of the opposite end.

The A-frame is challenging because the rather small feet of the robot do not provide enough friction for semi-static behaviors. The lack of friction also penalizes jittery motions that lose contact with the ground. Moreover, the sensitivity to friction and restitution of the surface and the feet deformations also make the problem harder to model perfectly in simulation. The A-frame requires the robot to approach with some speed and maintain good contact to generate sufficient friction while keeping  forward momentum. The robot must also keep its balance and transition to a soft landing during the downhill segment. 

\subsubsection{Broad Jump} The jump board is \SI{0.5}{\meter} long and \SI{1}{\meter} wide and sits flush with the floor. The robot has to clear it without touching. Jumping across a gap significantly exceeding the torso length is difficult for small quadruped robots to achieve from a static pose due to power/torque limits.

\subsubsection{Scoring}
The score is based on succeeded obstacles and completion time. Timing starts when the robot's center leaves the start table and stops when it reaches the end table.                               

\subsection{Barkour Experiment Setup}

\subsubsection{Barkour Operation \& Automation}

\begin{figure*}[t]
    \centering
    \includegraphics[width=1 \textwidth]{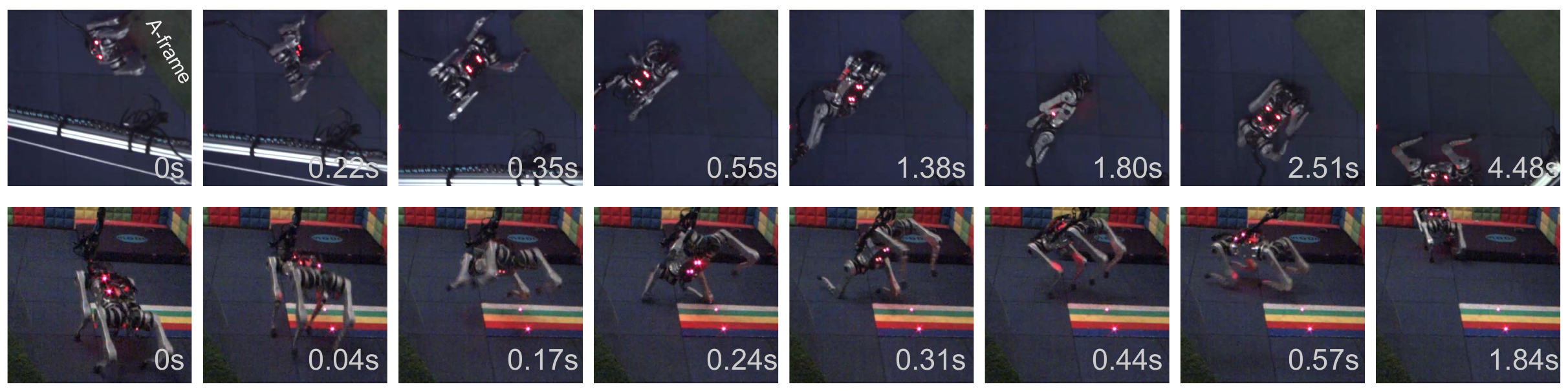}
    \caption{Examples of recovery behavior during Barkour runs with the Locomotion-Transformer policy. Top: Robot accidentally rolling over after descending the A-frame. This briefly triggers the recovery policy after which the robot continues towards the broad jump. Bottom: suboptimal takeoff during the broad jump causing the robot to land sideways with the knees hitting the floor. The policy recovers and continues towards the end table. 
    }
    \label{fig:robustness}
\end{figure*}

At the start of each run, the robot is positioned on top of the start table facing the second obstacle (the weave poles in this case). Every run is timed (down to at least \SI{0.01}{\second}) and the timer starts running when the center of the robot's torso leaves the borders of the start table. The timer stops running when the robot climbs on top of the end table. The trial is accepted as valid if the robot stands on the end table for \SI{5}{\second}.

To be able to restart an experiment without assistance, we incorporate a default policy for walking back to the start table and an automated recovery policy~\cite{smith2022legged} that triggers when the robot has tipped over. Fig.~\ref{fig:robustness} shows two typical examples of the robot successfully recovering from unexpected conditions. In the first example, the robot accidentally rolls over after completing the A-frame. This triggers a brief switch to the recovery policy, after which the Locomotion-Transformer policy repositions the robot towards the broad jump obstacle. In the second example, the robot suboptimally jumps into the air at the broad jump (likely due the cable pulling it at an inopportune time). After landing sideways with the knees hitting the floor, the generalist policy quickly recovers and the robot continues towards the end table.

\subsection{Distilling Hardware Data into a Course-Specific Locomotion-Transformer}
\label{appendix:hwdistill}
As indicated in the main text, training a Locomotion-Transformer policy based on expert policies in simulation raises a couple questions around the flexibility of the method:
\begin{itemize}
    \item \emph{Is the Transformer-based framework also capable of learning high-level behaviors}?
    \item \emph{Is it possible to train a Locomotion-Transformer policy from a small hardware dataset}? 
\end{itemize}

\begin{figure}
    \centering
    \includegraphics[width=0.48\textwidth]{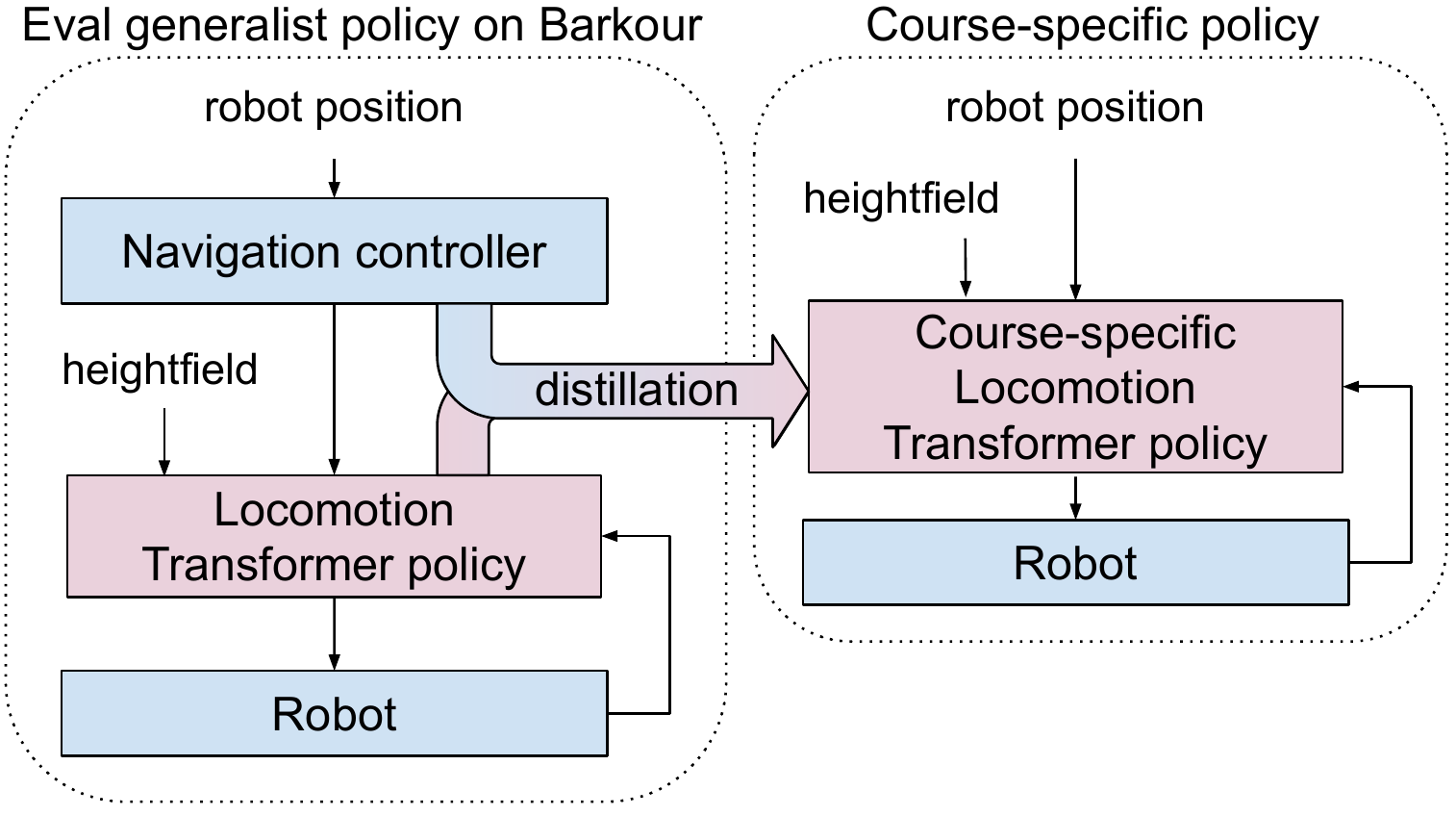}
    \caption{Training a course-specific Locomotion-Transformer policy based on data collected from a generalist Locomotion-Transformer with the Barkour navigation controller. We collect Barkour hardware episodes from the course using three expert policies in Fig.~\ref{fig:barkour_spec} and distill them into a new Locomotion-Transformer policy that is specific to this course and does not require a navigation controller. This shows that our framework can be used to learn agile skills as well as high-level (navigation) policies.}
    \label{fig:distill_all_the_things}
\end{figure}

To explore these topics, we collect a small dataset (64 episodes, 112K samples total) of Barkour runs that successfully completed the whole course from Fig.~\ref{fig:barkour_spec} on hardware using the navigation controller and specialist policies or generalist Locomotion-Transformer policies. 
Next, we use the distillation pipeline but remove the navigation controller and directly feed the robot's position and orientation to the Locomotion-Transformer (Fig.~\ref{fig:distill_all_the_things}).
In other words, instead of a generalist policy that accepts a broad range of commands from a navigation controller, we train a \emph{course-specific} policy.
More precisely, the course-specific policy directly maps the robot's position, orientation, proprioceptive sensors (IMU, joints), and heightfield to motor commands.

\begin{figure}
    \centering
    \includegraphics[width=0.3\textwidth]{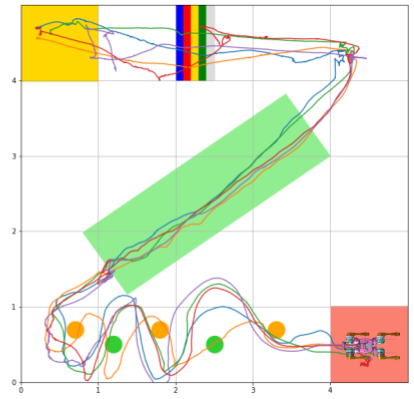}
    \caption{Typical trajectories for course-specific Locomotion-Transformer policy distilled based on only about 112K samples (64 episodes) of hardware data. During one episode (orange line) the robot missed a weave pole and the course-specific policy continued a correctly shaped but offset zig-zag pattern. The robot attempted the broad jump during all runs but doesn't fully clear it, indicating the complexity of the jump.}
    \label{fig:hw_distilled_trajectories}
\end{figure}

Remarkably, the course-specific policy is able to achieve good performance on the full Barkour benchmark and consistently completes the course (Fig.~\ref{fig:hw_distilled_trajectories}.
As the dataset is small, only about 80K iterations (about \SI{10}{\minute} of training time on a 4x4 TPUv3 cluster) were needed to achieve this level of performance. 
While in all runs the robot switches to a bounding gait and attempts the broad jump, it does not fully clear it, reflecting the complexity of this obstacle.
We also noticed that occasionally the policy skips or misses a weave pole, but still zig-zags as expected. This suggests that the policy has approximately learned the trajectory of the Barkour course as the weave poles are too small to be reliably sensed via the heightfield.
The runs shown in Fig.~\ref{fig:hw_distilled_trajectories} achieved a maximum agility score of $0.71$ and an average agility score of $0.57$, mainly lowered by incorrectly zig-zagging through the weave poles.
This result suggests the potential of extending the transformer-based learning pipeline beyond training low-level agile skills as demonstrated in the previous section and acquiring higher-level capabilities such as navigation.

\begin{figure}
    \centering
    \includegraphics[width=0.48\textwidth]{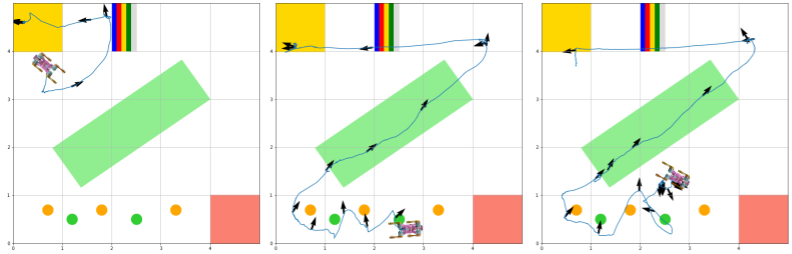}
    \caption{Robustness of course-specific Locomotion-Transformer policy. We placed the robot in a location or orientation near but outside the nominal Barkour trajectory and observed that the robot  moves back to the Barkour trajectory. The black arrows show the robot's orientation at regular time intervals.}
    \label{fig:hw_distilled_recovery}
\end{figure}

Finally, we qualitatively verified the robustness of course-specific Locomotion-Transformer policy by placing the robot near but not on the nominal trajectory for the course.
As shown in Fig.~\ref{fig:hw_distilled_recovery}, the policy is able walk towards the nominal trajectory and complete the course.
In some cases (Fig.~\ref{fig:hw_distilled_recovery} right plot, farther from nominal trajectory), the robot stands still until it receives a small push, after which it completes the course.
\subsection{OWP Observation Space and Training}
\label{app:owp_observation}
For the heightfield, OWP uses the same grid of $17 \times 11$ points measured around the robot, separated by \SI{0.1}{\meter}, and centered at the robot's CoM. During training, we randomly sample a new velocity command every \SI{10}{\second}. The sampling range is $\bar{v_x} \in [-1.5, 1.5]$\SI{}{\meter\per\second}, $\bar{v_y} \in [-1.0, 1.0]$\SI{}{\meter\per\second}, $\bar{\omega_z} \in [-\pi/2, \pi/2]$\SI{}{\radian\per\second}.

\subsection{SCP Terrain Curriculum and Training}
\label{app:scp_training}
The robot starts with the easiest terrain (\SI{5}{\degree}) and, if it is able to complete the slope, it advances to the next level (increment by 3 degrees). We adjust the sampling distribution of velocity commands during training of SCP to be $\bar{v_x} \in [-1.0, 1.5]$ \SI{}{\meter\per\second}, $\bar{v_y} \in [-0.5, 0.5]$ \SI{}{\meter\per\second}, and $\bar{\omega_z} \in [-0.5, 0.5]$ \SI{}{\radian\per\second}. 

\subsection{JP Training Curriculum}
\label{app:jp_training}
First the policy is trained on flat ground to bound forward quickly at a speed up to \SI{2.25}{\meter\per\second}. Next we introduce a series of gaps ranging from \SI{0.3}{\meter} to \SI{0.6}{\meter}. Finally, we fine-tune with a fixed \SI{0.6}{\meter} gap, randomized torque limits, and terminate the episodes if the robot's knees are less than \SI{0.06}{\meter} high over the gap. Detailed reward design are summarized in \autoref{tab:reward_terms}. The robot learns a bounding policy for this task, with symmetric left and right actions. To identify the broad jump when deployed to the robot, we mark the portion that the robot needs to jump over in the heightfield as $-1$, i.e., as if there is a gap.

\subsection{Details on Reward function for Specialist Policy Training}

\label{app:reward_function}

The specialist policy and value models consist of separate fully-connected MLPs each with four layers of sizes 1024, 512, 256, and 128. Each hidden layer is followed by an ELU activation and all inputs are flattened and fed directly into the network. 

\begin{table}[t]
\scriptsize
	\begin{center}
\sisetup{table-format=+1.4e+2, output-exponent-marker = \text{e} }
		\begin{tabular}{lllll}
			\toprule
                \vspace{0.1cm}
                \textbf{Reward term} & \textbf{Formulation} &  \multicolumn{3}{c}{\textbf{Weights per policy}}  \\
                & & \textbf{OWP} & \textbf{SDP} & \textbf{JP} \\
                \midrule
                Track Linear Velocity& $e^{-4(v_x-\bar{v_x})^2-4(v_y-\bar{v_y})^2}$ & $2$ & $0$ & $0$ \\
                
                Track Angular Velocity &
                $e^{-4(\bar{\omega_z} - \omega_z)^2}$ &
                $1$ & $0$ & $1$ \\
                
                Forward Velocity & $clip(v_x, -v_{\text{MAX}}, v_{\text{MAX}})$ & $0$ & $4$ & $3$\\
                
                Base Height & $-(p_z-\bar{p_z})^2$& $5$ & $0$ & $0$ \\
                
                Vertical Velocity & $-v_z^2$& $2$ & $2$ & $1$ \\
                
                Roll Pitch Velocity & $-\omega_x^2-\omega_y^2$ & $0.05$ & $0.05$ & $0.05$\\
                
                Orientation & $-g_z^2$ & $5$ & $0.5$ & $0.5$\\
                
                Torque & $-\tau^2$ & \num{2e-4} & \num{2e-4} & \num{2e-4} \\
                
                Joint Acceleration & $-\ddot{q}^2$ & $\num{2.5e-7}$  & \num{2.5e-7} & \num{2.5e-7} \\
                
                Action Delta &
                $-(a_{t}-a_{t-1})^2$ & $0.05$  & $0.05$ & $0.075$ \\
                
                Action Magnitude &
                $-(a_{t})^2$ & $0.02$ & $0.02$ & $0.025$ \\
                
                Feet Clearance & $\sum_{i=1}^{4} \mathbbm{1}_{f_{i} = 0} \cdot h_{i}$ & $0.25$ & $0.1$ & $0.1$ \\
                
                Stand Still & $\mathbbm{1}_{|\bar{v}| \leq 0.1} \cdot (q-\bar{q})^2$ & $0.1$ & $0$ & $0$ \\
                
                Jump Height & $p_z$ if upright over gap & $0$ & $0$ & $25$ \\
                
			\bottomrule
		\end{tabular}
	\end{center}
	\captionsetup{font={small,it},labelsep=colon}
	\caption{Reward terms used during training. $v_x$, $v_y$, $v_z$ are the base velocity in X, Y, Z axes, $\omega_z$ is the base yaw velocity, $p_z$ is the height of the base, $g$ is the projected gravity vector in robot space, $\tau$ is the applied torque on the motors, $\mathbbm{1}_{condition}$ is an indicator function that is evaluated to be $1$ if $condition$ is satisfied and $0$ otherwise, and $h_i$ denotes the height of the $i_{th}$ foot of the robot. \label{tab:reward_terms}}
\end{table}

\subsection{Details on Navigation Controller}
\label{appendix:Navigation Controller}
The navigation controller sends linear and angular velocity commands to the policy based on the next waypoint's relative position and target yaw angle:
$$ v_x = \alpha \sin(\phi)  \lVert \mathbf{d} \rVert,\, 
 v_y = \beta \cos(\phi) \lVert \mathbf{d} \rVert,\, \omega = \gamma \Delta(\psi), $$
where $\alpha$ and $\beta$ are coefficients for frontal linear velocity and sideways linear velocity, $\lVert \mathbf{d} \rVert$ is the distance between the center of the robot's torso and the next waypoint, $\phi$ is the angle between the vectors from the robot to the next waypoint and the robot's heading, $\gamma$ is the coefficient for angular velocity and $\Delta(\psi)$ is the difference between robot's heading angle and the waypoint's orientation. Lastly, $v_x$, $v_y$, and $\omega$ are clipped based on the policies' capabilities (depending on how fast policies are trained to walk and turn).

The active waypoint is only removed when it satisfies two constraints: 
    $\lVert \mathbf{d} \rVert < d_{lim}$ and 
    $\Delta(\psi) < \psi_{lim}$,
where $d_{lim}$ and $\psi_{lim}$ are thresholds defined per waypoint that depend on the desired precision. For example, while we set $d_{lim}$ to \SI{0.3}{\meter} and $\psi_{lim}$ to \SI{0.2}{\radian} for the weave poles, we lower them to \SI{0.1}{\meter} and \SI{0.17}{\radian} in front of a jump.

\subsection{Details on Locomotion-Transformer Data Collection}
\label{app:data_collection}

\begin{figure}[t]
    \centering
    \includegraphics[width=0.5\textwidth]{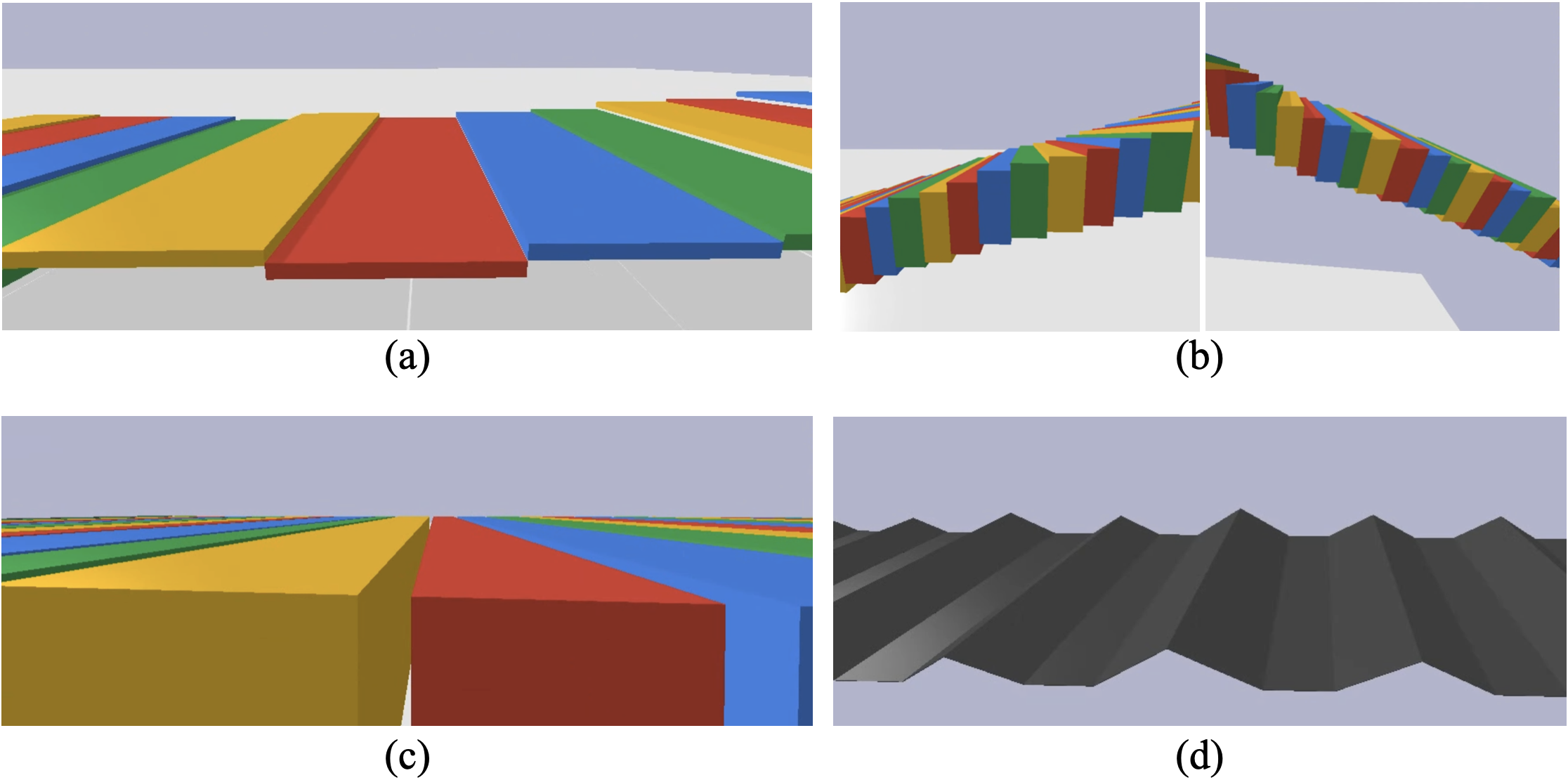}
    \caption{Data collection environments for training the generalist policy. (a) Random steps. (b) Stairs. (c) Gaps. (d) Slopes. \label{fig:data_collection_envs}}
\end{figure}

To train the generalist policy, we collect data in simulation by rolling out specialist policies.
\Cref{fig:data_collection_envs} illustrates the four environments that we use for data collection in this work.
To handle the observation space differences of skill policies, we merge the terrain heightfield grid so it simultaneously covers around the robot (for OWP) and also extends along the heading direction (for SDP and JP).
For policies that are not conditioned on an input speed command, we use the average realized velocity of the robot after \SI{1}{\second} as the effective command for a given moment.

To collect data of the robot climbing up/down steps and moving in different directions on uneven terrains, we use an environment with stairs and a random-step environment with step heights randomized between $[0.07, 0.16]$\SI{}{\meter} and deploy the OWP on them. 
For dashing up and down slopes, we design an environment consisting of a sequence of hills with slope inclinations in the $[0.5, 0.58]$\SI{}{\radian} range and use SDP to collect data. Finally, for the jumping skill we design a stage where the robot must jump over a series of gaps with widths sampled from $[0.5, 0.7]$\SI{}{\meter}. 
We implemented the data collection environments in PyBullet, though our framework is not specific to this design choice. 
We apply the same domain randomization scheme as during the training phase to obtain data near the optimal trajectories, which is important for sim-to-real transfer of the distilled policy. We collect $1,000$ episodes of data for stairs and random steps environments respectively, $8,000$ episodes for the gaps environment, and $5,000$ episodes for the slope environment. We terminate the data collection episode when the robot base pitch or roll angle is larger than \SI{70}{\degree}, i.e., when the robot loses its balance.

\begin{table}[ht]
\scriptsize
	\begin{center}
		\begin{tabular}{llll}
			\toprule
                \textbf{Environment} & Episodes & Steps & Robot Time (h) \\
                \midrule
                Jump & 10635 & 3940673 & 21.89 \\
                Flat & 1000 & 1495656 & 8.31 \\
                Slope & 3000 & 2336622 & 12.98 \\
                Stair Down & 1001 & 786536 & 4.37 \\
                Stair Up & 1000 & 1206722 & 6.70 \\
                Random Steps & 1000 & 598035 & 3.32 \\
                \midrule
                Total & 17636 & 10364244 & 57.58 \\
			\bottomrule
		\end{tabular}
	\end{center}
	\captionsetup{font={small,it},labelsep=colon}
	\caption{Locomotion-Transformer Training Data Card}
	\label{tab:transfurmer_data_card}
\end{table}

\subsection{Details on Locomotion-Transformer Architecture}
\label{app:transfurmer_architecture}

As described in \Cref{subsec:generalist}, Locomotion-Transformer is a causal Transformer~\cite{vaswani2017attention}, following the standard GPT implementation~\cite{radford2019language}.
We have two layers and the context length is 15.
The transformer input token and the internal representations are 256-$d$. 
For tokenizing elevation maps, we use one convolutional encoder to map each elevation map into a 64-$d$ embedding vector, concatenate all embedding vectors, and project the concatenated vector into a 256-$d$ token.
All the convolutional encoders have two convolution layers with kernel size 3, channel size 8, and stride size 1, followed by a 64$-d$ projection layer.
Proprioceptive states and velocity commands are tokenized together with a 256-$d$ projection following a concatenation.
Actions are also tokenized with a 256-$d$ projection.
All tokenizer hidden layers are followed by ReLU activation.

\subsection{Accelerating Training with Brax}
\label{appendix:brax}

In parallel to our efforts to train specialist policies in IsaacGym, we also investigated increasing iteration speed by training specialist policies on 4x2 TPU clusters with Brax~\cite{freeman2021brax}. In general, we are able to reach similar reward values and correspondingly similar in-sim performance in Brax environments within about \SI{2}{\minute}, after 200 million simulation steps using Brax's JAX-based PPO implementation. As Brax is a comparatively newer engine, some features also had to be developed specifically for this work, including domain randomization and custom collision primitives.

\subsubsection{System Identification}
Because of the difference in implementation details between the various physics engines used, we designed a simple protocol to align the actuation parameters between the Brax robot model and the other (sim and real) models. We placed the robot on a raised platform and recorded a trace of its joint angles when actuating to a target position from a reference neutral pose. We then differentiably optimized the joint gain and damping parameters until the joint trajectories had converged.  

\subsubsection{Training Environments}
\label{subsec:braxtrainingenv}

We followed the environment setup outlined in Sec.~\ref{sec:control_framework}, sharing the observation space, actuation model, reward functions, and domain randomization scheme. However, we did not use a curriculum, and instead uniformly sampled over environment variability. 
\begin{figure}[ht]
    \centering
    \includegraphics[scale=0.35]{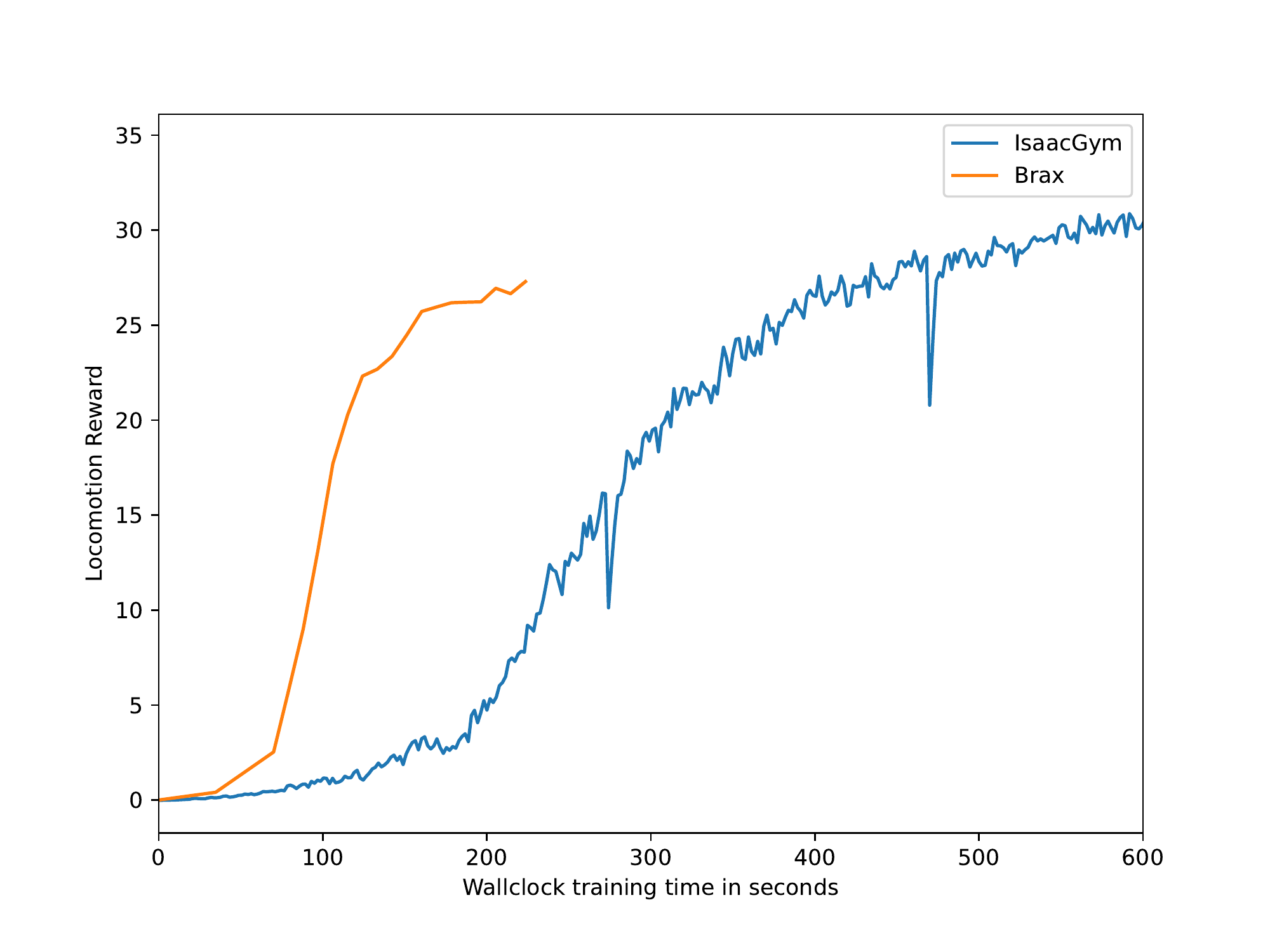}
    \caption{Representative training curves of Brax vs. IsaacGym.  After \SI{200}{\second}, Brax has simulated approximately 200 million environment steps, and after \SI{600}{\second}, IsaacGym has simulated approximately 32 million environment steps.  Both training runs used the same reward function parameters for an omnidirectional walking policy (i.e., column 1. in Table~\ref{tab:reward_terms}).}
    \label{fig:brax_v_isaac}
\end{figure}

We provide a reference training curve of the omnidirectional walking policy in Fig.~\ref{fig:brax_v_isaac} and validated the resulting policies on hardware.
\begin{figure}[ht]
    \centering
    \includegraphics[width=0.45\textwidth]{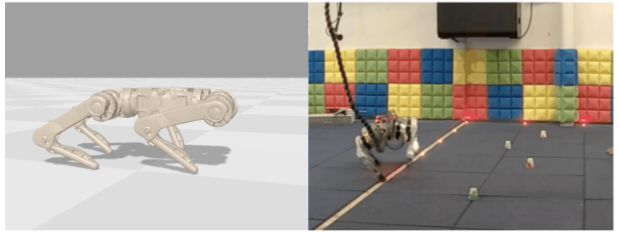}
    \caption{Snapshots of forward locomotion in (left): the Brax simulator, and (right): the policy being deployed to the robot.}
    \label{fig:brax_bullet_real}
\end{figure}

For the A-Frame environment, we trained policies for traversing the A-Frame as well as sloped terrain variants for different slope angle/friction combinations. The behavior of these policies tended to depend sensitively on the distribution of slopes and frictions being randomized over. 
These sloped-traversal policies typically performed well in sim, but had trouble transferring to the real, rugged A-frame obstacle when deployed to the robot.

We explored additional variants of locomotion rewards, including the ``Fetch'' reward from Brax's default set of environments~\cite{freeman2021brax}, as well as various penalization and reward schemes for encouraging more graceful motions. Notably, Fetch worked out of the box to produce a gait capable of locomotion (without domain randomization), but the motion quality and strain on the robot were too jittery for practical use.  

We also explored alternative environment definitions, including different observation space features, and different bread-crumbing mechanisms for the target, but ultimately settled on the scheme described in the main text.

\subsubsection{Validation}
\label{subsec:braxvalidation}
We evaluated cross-engine correspondence fidelity by qualitatively verifying IsaacGym policies running in Brax, Brax-learned policies running in PyBullet, and Brax policies running on the robot (see, e.g., Fig.~\ref{fig:brax_bullet_real}).
Reaching qualitatively reasonable transfer quality required domain randomization as well as motor-level gain and damping coefficient optimization.  We defer a more thorough analysis of sim-to-real transfer between Brax and the robot to future work. At a high level, more aggressive domain randomization and larger observation noise typically resulted in better transfer at the expense of more conservative gaits.

\subsubsection{Conclusions and Future Work}
\label{subsec:brax_conclusions}

\begin{figure}[ht]
    \centering
    \includegraphics[scale=0.35]{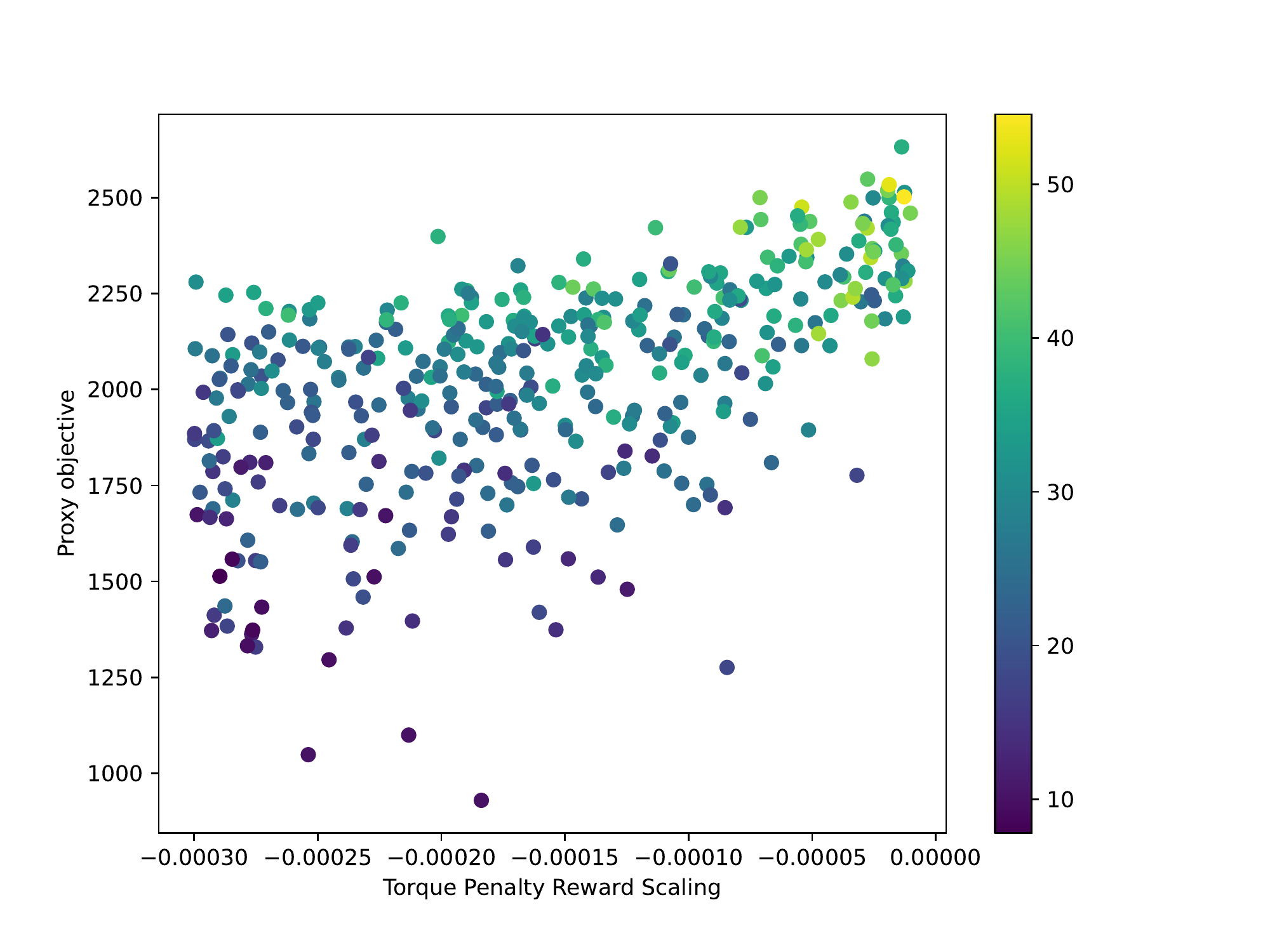}
    \caption{A representative hyperparameter sweep over the LeggedGym reward parameters, centered on the default parameters. Each point represents a different random sampling of parameters, with the Torque penalty on the x-axis (see "Torque" in Table.~\ref{tab:reward_terms}), and a proxy objective (the integrated average distance a policy reached, averaged over 128 random episodes) on the y-axis.  In color, we plot the raw reward value which is unsurprisingly largest (farthest distance traveled) for the least penalized policies.}
    \label{fig:brax_hyper_sweep}
\end{figure}

Brax proved to be a fast iteration platform, supporting near-interactive reward shaping experiments and seamless scaling to large numbers of accelerators. In particular, we note Brax's efficiency to facilitate high-throughput hyperparameter sweeps for effective reward scaling coefficient settings (Fig.~\ref{fig:brax_hyper_sweep}).  
\balance
The primary obstacles to its uptake were largely related to development velocity, feature completeness, and motion quality: existing, functioning pipelines needed to be duplicated within Brax, Brax did not always have the necessary features out-of-the-box that already existed in, e.g., IsaacGym, and Brax policies simply weren't as smooth, owing to less overall researcher-time spent refining them. Nonetheless, Brax remains a compelling platform for sim-to-real robotics research.
The explorations described above were based on Brax v1 as the next generation of Brax (v2) was not yet available at the time of writing.

\end{document}